\documentclass[11pt, oneside]{Thesis}

\usepackage{siunitx}
\usepackage[square, numbers, comma, sort&compress]{natbib} 
\usepackage{mathtools}
\hypersetup{urlcolor=blue, colorlinks=true} 
\title{\ttitle} 
\usepackage[export]{adjustbox}
\usepackage{xcolor,colortbl}
\usepackage{url}
\usepackage{subcaption}
\usepackage{array}
\definecolor{darkgray}{rgb}{.4,.4,.4}

\definecolor{purple}{rgb}{0.65, 0.12, 0.82}

\usepackage{afterpage}

\newcommand\blankpage{%
    \null
    \thispagestyle{empty}%
    \addtocounter{page}{-1}%
    \newpage}
    
\begin{document}

\frontmatter 

\setstretch{1.3}

\fancyhead{} 
\rhead{\thepage} 
\lhead{} 

\pagestyle{fancy} 

\newcommand{\HRule}{\rule{\linewidth}{0.5mm}} 

\hypersetup{pdftitle={\ttitle}}
\hypersetup{pdfsubject=\subjectname}
\hypersetup{pdfauthor=\authornames}
\hypersetup{pdfkeywords=\keywordnames}


\begin{titlepage}
\begin{center}

\includegraphics[width=15cm, height=4cm]{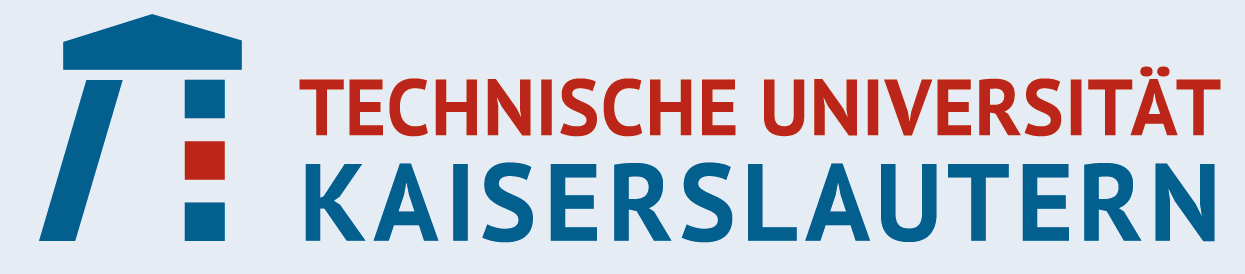}

\textsc{\LARGE \univname}\\[1.5cm] 
\textsc{\Large Master Thesis}\\[0.5cm] 

\HRule \\[0.4cm] 
{\huge \bfseries \ttitle}\\[0.4cm] 
\HRule \\[1.5cm] 
 
\begin{minipage}{0.4\textwidth}
\begin{flushleft} \large
\emph{Author:}\\
{\authornames} 
\end{flushleft}
\end{minipage}
\begin{minipage}{0.4\textwidth}

\begin{flushright} \large
\emph{Supervisors:} \\
\href{http://www3.dfki.uni-kl.de/agd/dengel/content/index_ger.html}{\pname} 
\end{flushright}

\begin{flushright} \large
\href{http://www.dfki.uni-kl.de/~cschulze/}{\mname} 
\end{flushright}

\end{minipage}\\[2cm]
 
\large \textit{A thesis submitted in fulfilment of the requirements\\ for the \degreename}\\[0.3cm] 
\textit{in the}\\[0.4cm]
\deptname\\[1cm] 
 
{\large January, 2016}\\[2cm]

\end{center}

\end{titlepage}

\afterpage{\blankpage}

\clearpage 

\Declaration{

\addtocontents{toc}{\vspace{1em}} 

I, \authornames, declare that this thesis titled, `\ttitle' and the work presented in it are my own. I confirm that:

\begin{itemize} 
\item[\tiny{$\blacksquare$}] This work was done mainly while in candidature for a masters degree at this University.
\item[\tiny{$\blacksquare$}] Where I have consulted the published work of others, this is always clearly attributed.
\item[\tiny{$\blacksquare$}] Where I have quoted from the work of others, the source is always given. With the exception of such quotations, this thesis is entirely my own work.
\item[\tiny{$\blacksquare$}] I have acknowledged all main sources of help.
\end{itemize}
 
Signature:\\
\rule[1em]{25em}{0.5pt} 
 
Date:\\
\rule[1em]{25em}{0.5pt}
}

\afterpage{\blankpage}

\clearpage 


\pagestyle{empty} 

\null\vfill 

\textit{``Satisfaction lies in the effort, not in the attainment, full effort is full victory."}

\begin{flushright}
Mahatma Gandhi
\end{flushright}

\vfill\vfill\vfill\vfill\vfill\vfill\null 

\afterpage{\blankpage}

\clearpage


\addtotoc{Abstract} 

\abstract{\addtocontents{toc}{\vspace{1em}} 

In the recent years, there has been a tremendous increase in the amount of video content uploaded to social networking and video sharing websites like Facebook and Youtube. As of result of this, the risk of children getting 
exposed to adult and violent content on the web also increased. To address this issue, an approach to automatically detect violent content in videos is proposed in this work. Here, a novel attempt is made also to detect the 
category of violence present in a video. A system which can automatically detect violence from both Hollywood movies and videos from the web is extremely useful not only in parental control but also for applications 
related to movie ratings, video surveillance, genre classification and so on. 

Here, both audio and visual features are used to detect violence. MFCC features are used as audio cues. Blood, Motion, and SentiBank features are used as visual cues. Binary SVM classifiers are trained on each of these 
features to detect violence. Late fusion using a weighted sum of classification scores is performed to get final classification scores for each of the violence class target by the system. To determine optimal weights for
each of the violence classes an approach based on grid search is employed. Publicly available datasets, mainly Violent Scene Detection (VSD), are used for classifier training, weight calculation, and testing. The performance
of the system is evaluated on two classification tasks, Multi-Class classification, and Binary Classification. The results obtained for Binary Classification are better than the baseline results from MediaEval-2014.

}

\afterpage{\blankpage}

\clearpage


\setstretch{1.3} 

\acknowledgements{\addtocontents{toc}{\vspace{1em}} 
\begin{itemize} 
\item First of all, I would like to express my gratitude to my supervisor Dr. Christian Schulze for his support.

\item I would like to thank my professors Prof. Andreas Dengel for introducing me to this topic.

\item Furthermore, I would like to thank my loved ones, who have supported me through out my journey. 

\item I will be grateful to you all, for all your care and support.
\end{itemize}

}
\afterpage{\blankpage}
\clearpage 


\pagestyle{fancy} 

\lhead{\emph{Contents}} 
\tableofcontents 

\lhead{\emph{List of Figures}} 
\listoffigures 

\lhead{\emph{List of Tables}} 
\listoftables 



\clearpage 

\setstretch{1.5} 

\lhead{\emph{Abbreviations}} 
\listofsymbols{ll} 
{
\textbf{ANP} & \textbf{A}djective \textbf{N}oun \textbf{P}air \\
\textbf{AP} & \textbf{A}verage \textbf{P}recision \\
\textbf{BPM} & \textbf{B}lood \textbf{P}robability \textbf{M}ap \\
\textbf{EER} & \textbf{E}qual \textbf{E}rror \textbf{R}ate \\
\textbf{HSV} & \textbf{H}ue \textbf{S}aturation \textbf{V}alue \\
\textbf{MAP} & \textbf{M}ean \textbf{A}verage \textbf{P}recision \\
\textbf{MFCC} & \textbf{M}el \textbf{F}requency \textbf{C}epstral \textbf{C}oefficients\\
\textbf{MoSIFT} & \textbf{Mo}tion \textbf{S}cale \textbf{I}nvariant \textbf{F}eature \textbf{T}ransform \\
\textbf{RBF} & \textbf{R}adial \textbf{B}asis \textbf{F}unction \\
\textbf{RGB} & \textbf{R}ed \textbf{G}reen \textbf{B}lue \\
\textbf{ROC} & \textbf{R}eceiver \textbf{O}perating \textbf{C}haracteristic \\
\textbf{SIFT} & \textbf{S}cale \textbf{I}nvariant \textbf{F}eature \textbf{T}ransform \\
\textbf{STIP} & \textbf{S}pace \textbf{T}ime \textbf{I}nterest \textbf{P}oints \\
\textbf{SVM} & \textbf{S}upport \textbf{V}ector \textbf{M}achines \\
\textbf{ViF} & \textbf{Vi}olent \textbf{F}lows \\
\textbf{VSD} & \textbf{V}ioent \textbf{S}cene \textbf{D}ataset \\
\textbf{VSO} & \textbf{V}isual \textbf{S}entiment \textbf{O}ntology \\
\textbf{XML} & E\textbf{X}tended \textbf{M}arkup \textbf{L}anguage \\
\textbf{ZCR} & \textbf{Z}ero \textbf{C}ross \textbf{R}ate \\
}

\afterpage{\blankpage}
\clearpage


\setstretch{1.3} 

\pagestyle{empty} 

\dedicatory{This thesis is dedicated to my father, for his love, support, and encouragement\ldots} 

\addtocontents{toc}{\vspace{2em}} 

\afterpage{\blankpage}
\clearpage


\mainmatter 

\pagestyle{fancy} 


\chapter{Introduction} 

\label{Chapter1}  

\lhead{Chapter 1. \emph{Introduction}} 

The amount of multimedia content uploaded to social networking websites and the ease with which these can be accessed by children is posing a problem to parents who wish to protect their children from getting exposed 
to violent and adult content on the web. The number of video uploads to websites like YouTube and Facebook are on the rise. There is an increase of \num{75}\% in the number of video posts on Facebook (\citet{facebookstats}) in
the last one year and more than \num{120000} videos are uploaded to YouTube every day (\citet{youtubestats}, \citet{gill2007}). It is estimated that \num{20}\% of the videos uploaded to these websites
contain violent or adult content (\citet{sparks2015}). This makes it easy for children to access or accidentally get exposed to these unsafe contents. The effects of watching violent content on children are well studied in 
psychology (\citet{tompkins2003}, \citet{sparks2015}, \citet{bushman2006}, and \citet{huesmann2006}) and  the results of these studies suggest that watching of violent content has a substantial effect on emotions of the
children. The major effects are increases in the likelihood of aggressive or fearful behavior and becoming less sensitive to the pain and suffering of others. \citet{huesmann2013} conducted a study involving children from 
elementary school, who watched many hours of violence on television. By observing these children into adulthood, they found that the ones who did watch a lot of television violence when they were \num{8} years old were more
likely to be arrested and prosecuted for criminal acts as adults. Similar studies by \citet{flood2009} and \citet{mitchell2003} suggest that exposure to adult content also has detrimental effects on children. This motivated 
research in the field of automatic violent and adult content detection in videos. 

Adult content detection (\citet{chan1999}, \citet{schulze2014}, \citet{pogrebnyak2015}) is well studied and much progress has been made. Violence detection, on the other hand, has been less studied and has gained interest only
in the recent past. Few approaches for violence detection were proposed in the past and each of these approaches tried to detect violence using different visual and auditory features. For example, \citet{nam1998} combined
multiple audio-visual features to identify violent scenes. In their work, flames and blood were detected using predefined color tables and various representative audio effects (gunshots, explosions, etc.) were also exploited. 
\citet{datta2002} proposed an accelerated motion vector based approach to detect human violence such as fist fighting, kicking, etc. \citet{cheng2003} presented a hierarchical approach to locating gun play and car racing 
scenes through detection of typical audio events (e.g. gunshots, explosions, and car-braking). 

More approaches proposed for violence detection are discussed in \cref{Chapter2}. All of these approaches focused mainly only on detection of violence in Hollywood movies but not in videos from video sharing and social
media websites such as YouTube or Facebook. Detection of violence in Hollywood movies is relatively easy as these movies follow some moviemaking rules. For example, to exhibit exciting action scenes, the atmosphere of fast-pace 
is created through high-speed visual movement and fast-paced sound. But the videos from the video-sharing websites, like YouTube and Facebook, do not follow these moviemaking rules and often have poor audio and video quality.
These characteristics of user-generated videos make it very hard to detect violence in them.  

Before the approach to detect violence is discussed, it is important to provide a definition for the term ``Violence". All of the previous approaches for violence detection have not followed the same definition of violence and 
have used different features and different datasets. This makes the comparison of different approaches very difficult. To overcome this problem and to foster research in this area, a dataset named Violent Scene Detection
(VSD) was introduced by \citet{demarty2011} in 2011 and the recent version of this dataset is the VSD2014. According to this latest dataset, ``Violence" in a video is, ``any scene one would not let an \num{8} year old child
watch because they contain physical violence"\citet{schedl6vsd2014}. This definition is believed to be formulated based on the research findings from psychology, which are mentioned above. From this definition, it can be
observed that violence is not a physical entity but a concept which is very generic, abstract and also very subjective. Hence, violence detection is not a trivial task.

The aim of this work is to build a system which automatically detects violence not only in Hollywood movies, but also in videos from the video-sharing websites like YouTube and Facebook. In this work, an attempt is made to also 
detect the category of violence in a video, which was not addressed by earlier approaches. The categories of violence which are targeted in this work are the presence of blood, presence of cold arms, explosions, 
fights, screams, presence of fire, firearms, and gunshots. These represent the subset of concepts defined and used in the VSD2014 for annotating video segments. The categories ``gory scenes" and ``car chase" from VSD2014 were
not selected as there were not many video segments in VSD2014 annotated with these concepts. Another such category is the ``Subjective Violence". It is not selected as the scenes belonging to this category do not have any 
visible violence and hence are very hard to detect. In this work, both audio and visual features are used for violence detection as combining both audio and visual information provides more reliable results in classification.

The advantages of developing a system like this, which can automatically detect violence in multi-media content are many. It can be used to rate movies depending on the amount of violence. This can be used by 
social networking sites to detect and block upload of violent videos to their platforms. Also, it can be used for scene characterization and genre classification which helps in searching and browsing movies. Recognition of 
violence in video streams from real-time camera systems will be very helpful for video surveillance in places such as airports, hospitals, shopping malls, public places, prisons, psychiatric wards, school playgrounds etc.
However, real time detection of violence is much more difficult and in this work no attempt is made to deal with it.

An overview of related work, detailed description of the proposed approach and the evaluation are presented next. The following chapters are organized as follows. In \cref{Chapter2} some of the previous works in the area of 
violence detection are explained in detail. In \cref{Chapter3}, the details of the approach used for training and testing of feature classifiers are presented. It also includes the details of feature extraction and the 
classifier training. \cref{Chapter4} describes the details of datasets used, experimental setup and the results obtained from the experiments. Finally, in \cref{Chapter5} conclusions are provided followed by the 
possible future work.

\afterpage{\blankpage}
\clearpage

\chapter{Related Work} 

\label{Chapter2} 

\lhead{Chapter 2. \emph{Related Work}} 

Violence Detection is a sub-task of activity recognition where violent activities are to be detected from a video. It can also be considered as a kind of multimedia event detection. Some approaches have 
already been proposed to address this problem. These proposed approaches can be classified into three categories: (i) Approaches in which only the visual features are used. (ii) Approaches in which only the audio features
are used. (iii) Approaches in which both the audio and visual features are used. The category of interest here is the third one, where both video and audio are used. This chapter provides an overview of some of the previous
approaches belonging to each of these categories. 

\section{Using Audio and Video}

The initial attempt to detect violence using both audio and visual cues is by \citet{nam1998}. In their work, both the audio and visual features are exploited to detect violent scenes and generate indexes 
so as to allow for content-based searching of videos. Here, the spatio-temporal dynamic activity signature is extracted for each shot to categorize it to be violent or non-violent. This spatio-temporal dynamic activity 
feature is based on the amount of dynamic motion that is present in the shot. 

The more the spatial motion between the frames in the shot, the more significant is the feature. The reasoning behind this approach is that most of the action scenes involve a rapid and significant amount of movement of 
people or objects. In order to calculate the spatio-temporal activity feature for a shot, motion sequences from the shot are obtained and are normalized by the length of the shot to make sure that only the shots with 
shorter lengths and high spatial motion between the frames have higher value of the activity feature.  

Apart from this, to detect flames from gunshots or explosions,  a sudden variation in intensity values of the pixels between frames is examined. To eliminate false positives, such as intensity variation because of camera 
flashlights, a pre-defined color table with color values close to the flame colors such as yellow, orange and red are used. Similarly to detect blood, which is common in most of the violent scenes, pixel colors within a
frame are matched with a pre-defined color table containing blood-like colors. These visual features by itself are not enough to detect violence effectively. Hence, audio features are also considered. 

The sudden change in the energy level of the audio signal is used as an audio cue. The energy entropy is calculated for each frame and the sudden change in this value is used to identify violent events such as explosion or 
gunshots. The audio and visual clues are time synchronized to obtain shots containing violence with higher accuracy. One of the main contributions of this paper is to highlight the need of both audio and visual cues to 
detect violence. 

\citet{gong2008} also used both visual and audio cues to detect violence in movies. A three-stage approach to detect violence is described. In the first stage, low-level visual and auditory features are extracted
for each shot in the video. These features are used to train a classifier to detect candidate shots with potential violent content. In the next stage, high-level audio effects are used to detect candidate shots. In 
this stage, to detect high-level audio effects, SVM classifiers are trained for each category of the audio effect by using low-level audio features such as power spectrum, pitch, MFCC (Mel-Frequency Cepstral Coefficients) 
and harmonicity prominence (\citet{cai2006}). The output of each of the SVMs can be interpreted as probability mapping to a sigmoid, which is a continuous value between [\num{0},\num{1}] (\citet{platt1999}). 
In the last stage, the probabilistic outputs of first two stages are combined using boosting and the final violence score for a shot is calculated as a weighted sum of the scores from the first two stages. 

These weights are calculated using a validation dataset and are expected to maximize the average precision. The work by \citet{gong2008} concentrates only on detecting violence in movies where universal film-making rules are
followed. For instance, the fast-paced sound during action scenes. Violent content is identified by detecting fast-paced scenes and audio events associated with violence such as explosions and gunshots. The training and
testing data used  are from a collection of four Hollywood action movies which contain many violent scenes. Even though this approach produced good results it should be noted that it is optimized to detect violence only in
movies which follow some film-making rules and it will not work with the videos that are uploaded by the users to the websites such as Facebook, Youtube, etc.

In the work by \citet{lin2009}, a video sequence is divided into shots and for each shot both the audio and video features in it are classified to be violent or non-violent and the outputs are combined using 
co-training. A modified pLSA algorithm (\citet{hofmann2001}) is used to detect violence from the audio segment. The audio segment is split into audio clips of one second each and is represented by a feature vector containing  
low-level features such as power spectrum, MFCC, pitch, Zero Cross Rate (ZCR) ratio and harmonicity prominence (\citet{cai2006}). These vectors are clustered to get cluster centers which denote an audio vocabulary. Then, each 
audio segment is represented using this vocabulary as an audio document. The Expectation Maximization algorithm (\citet{dempster1977}) is used to fit an audio model which is later used for classification of audio segments. 
To detect violence in a video segment, the three common visual violent events: motion, flame/explosions and blood are used. Motion intensity is used to detect areas with fast motion and to extract motion features for each 
frame, which is then used to classify a frame to be violent or non-violent. Color models and motion models are used to detect flame and explosions in a frame and to classify them. Similarly, color model and motion intensity are
used to detect the region containing blood and if it is greater than a pre-defined value for a frame, it is classified to be violent. The final violence score for the video segment is obtained by the weighted sum of the three
individual scores mentioned above. The features used here are same as the ones used by \citet{nam1998}. For combining the classification scores from the video and the audio stream, co-training is used. For training and testing,
a dataset consisting of five Hollywood movies is used and precision of around \num{0.85} and recall of around \num{0.90} are obtained in detecting violent scenes. Even this work targets violence detection only in movies but 
not in the videos available on the web. But the results suggest that the visual features such are motion and blood are very crucial for violence detection.

\section{Using Audio or Video}
All the approaches mentioned so far use both audio and visual cues, but there are others which used either video or audio to detect violence and some others which try to detect only one a specific kind of violence such as
fist fights. A brief overview of these approaches is presented next. 

One of the only works which used audio alone to detect semantic context in videos is by \citet{cheng2003}, where a hierarchical approach based on Gaussian mixture models and Hidden Markov models is used to recognize
gunshots, explosions, and car-braking. \citet{datta2002} tried to detect person-on-person violence in videos which involve only fist fighting, kicking, hitting with objects etc., by analyzing violence at 
object level rather than at the scene level as most approaches do. Here, the moving objects in a scene are detected and a person model is used to detect only the objects which represent persons. From this, the motion trajectory
and orientation information of a person's limbs are used to detect person-on-person fights.

\citet{clarin2005} developed an automated system named DOVE to detect violence in motion pictures. Here, blood alone is used to detect violent scenes. The system extracts key frames from each scene and passes them to a trained 
Self-Organizing Map for labeling the pixels with the labels: skin, blood or nonskin/nonblood. Labeled pixels are then grouped together through connected components and are observed for possible violence. A scene is considered to 
be violent if there is a huge change in the pixel regions with skin and blood components. One other work on fight detection is by \citet{nievas2011} in which Bag-of-Words framework is used along with the action descriptors 
Space-Time Interest Points (STIP - \citet{laptev2005}) and Motion Scale-invariant feature transform (MoSIFT - \citet{chen2009}). The authors introduced a new video dataset consisting of \num{1000} videos, divided into two groups
fights and non-fights. Each group has \num{500} videos and each video has a duration of one second. Experimentation with this dataset has produced a \num{90}\% accuracy on a dataset with fights from action movies.

\citet{deniz2014} proposed a novel method to detect violence in videos using extreme acceleration patterns as the main feature. This method is \num{15} times faster than the state-of-the-art action recognition systems and also
have  very  high accuracy in detecting scenes containing fights. This approach is very useful in real-time violence detection systems, where not only accuracy but also speed matters. This approach compares the power 
spectrum of two consecutive frames to detect sudden motion and depending on the amount of motion, a scene is classified to be violent or non-violent. This method does not use feature tracking to detect motion, which 
makes it immune to blurring. \citet{hassner2012} introduced an approach for real-time detection of violence in crowded scenes. This method considers the change of flow-vector magnitudes over time. These changes 
for short frame sequences are called Violent Flows (ViF) descriptors. These descriptors are then used to classify violent and non-violent scenes using a linear Support Vector Machine (SVM). As this method uses only flow 
information between frames and forgo high-level shape and motion analysis, it is capable of operating in real-time. For this work, the authors created their own dataset by downloading videos containing violent crowd
behavior from Youtube.

All these works use different approaches to detect violence from videos and all of them use their own datasets for training and testing. They all have their own definition of violence. This demonstrates a major
problem for violence detection, which is the lack of independent baseline datasets and a common definition of violence, without which the comparison between different approaches is meaningless. 

To address this problem, \citet{demarty2012} presented a benchmark for automatic detection of violence segments in movies as part of the multimedia benchmarking initiative MediaEval-2011 \footnote{http://www.multimediaeval.org}. 
This benchmark is very useful as it provides a consistent and substantial dataset with a common definition of violence and evaluation protocols and metrics. The details of the provided dataset are discussed in detail in 
\sref{Datasets}. Recent works on violence recognition in videos have used this dataset and details about some of them are provided next.

\section{Using MediaEval VSD}

\citet{acar2011} proposed an approach that merges visual and audio features in a supervised manner using one-class and two-class SVMs for violence detection in movies. Low-level visual and audio features are extracted
from video shots of the movies and then combined in an early fusion manner to train SVMs. MFCC features are extracted to describe the audio content and SIFT (Scale-Invariant Feature Transform - \citet{lowe2004}) based
Bag-of-Words approach is used for visual content. 

\citet{jiang2012} proposed a method to detect violence based on a set of features derived from the appearance and motion of local patch trajectories(\citet{jiang2012_1}). Along with these patch trajectories, other 
features such as SIFT, STIP, and MFCC features are extracted and are used to train an SVM classifier to detect different categories of violence. Score and feature smoothing are performed to increase the accuracy. 

\citet{lam2013} evaluated the performance of low-level audio/visual features for the violent scene detection task using the datasets and evaluation protocols provided by MediaEval. In this work both the local and global
visual features are used along with motion and MFCC audio features. All these features are extracted for each keyframe in a shot and are pooled to form a single feature vector for that shot. An SVM classifier is 
trained to classify the shots to be violent or non-violent based on this feature vector. \citet{eyben2013} applied large-scale segmental feature extraction along with audio-visual classification for detecting violence.
The audio feature extraction is done with the open-source feature extraction toolkit openSmile(\citet{eyben2015}). Low-level visual features such as Hue-Saturation-Value (HSV) histogram, optical flow analysis, and Laplacian
edge detection are computed and used for violence detection. Linear SVM classifiers are used for classification and a simple score averaging is used for fusion.

\section{Summary}
In summary, almost all methods described above try to detect violence in movies using different audio and visual features with an expectation of only a couple [\citet{nievas2011}, \citet{hassner2012}], which use video
data from surveillance cameras or from other real-time videos systems. It can also be observed that not all these works use the same dataset and each have their own definition of violence. The introduction of the
MediaEval dataset for Violent Scene Detection (VSD) in 2011, has solved this problem. The recent version of the dataset, VSD2014 also includes video content from Youtube apart from the Hollywood movies 
and encourages researchers to test their approach on user-generated video content. 

\section{Contributions}

The proposed approach presented in \cref{Chapter3} is motivated by the earlier works on violence detection, discussed in \cref{Chapter2}. In the proposed approach, both audio and visual cues are used to detect violence.
MFCC features are used to describe audio content and blood, motion and SentiBank features are used to describe video content. SVM classifiers are used to classify each of these features and late fusion is applied to fuse the
classifier scores.

Even though this approach is based on earlier works on violence detection, the important contributions of it are: (i) Detection of different classes of violence. Earlier works on violence detection concentrated only on 
detecting the presence of violence in a video. This proposed approach is one of the first to tackle this problem. (ii) Use of SentiBank feature to describe visual content of a video. SentiBank is a visual feature which is used
to describe the sentiments in an image. This feature was earlier used to detect adult content in videos (\citet{schulze2014}). In this work, it is used for the first time to detect violent content. (iii) Use of 3-dimensional
color model, generated using images from the web, to detect pixels representing blood. This color model is very robust and has shown very good results in detecting blood. (iv) Use of information embedded in a video codec to 
generate motion features. This approach is very fast when compared to the others, as the motion vectors for each pixel are precomputed and stored in the video codec. A detailed explanation of this proposed approach is presented
in the next chapter, \cref{Chapter3}. 

\chapter{Proposed Approach} 

\label{Chapter3} 

\lhead{Chapter 3. \emph{Proposed Approach}} 

\begin{figure}[htbp]
\captionsetup{justification=justified,
singlelinecheck=false
}
\centering
\includegraphics[scale=0.5]{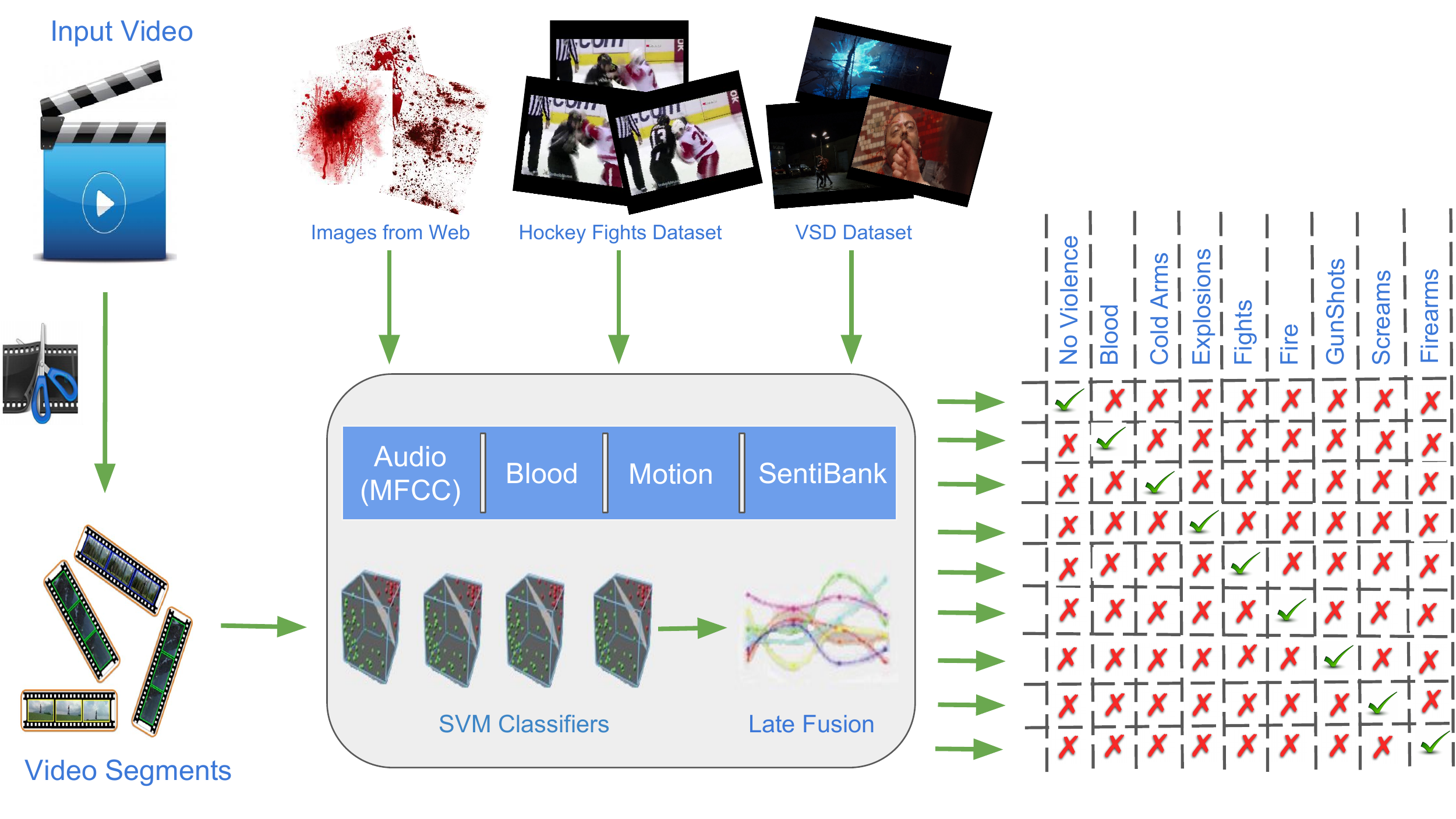}
\rule{\textwidth}{0.5pt}
\caption[Figure showing the overview of the system.]{Figure showing the overview of the system. Four different SVM classifiers are trained, one each for Audio, Blood, Motion and SentiBank features. Images from the web are used
 to develop a blood model to detect blood in  video frames. To train classifiers for all the features, data from the VSD2104 dataset is used. Each of these classifiers individually give the probability of a video segment 
 containing violence. These individual probabilities are then combined using the late fusion technique and the final output probability, which is the weighted sum of individual probabilities, is presented as output by the 
 system. The video provided as input to the system is divided into one-second segments and the probability of each of the segments containing violence is obtained as the output.
}
\label{fig:overview}
\end{figure}

This chapter provides a detailed description of the approach followed in this work. The proposed approach consists of two main phases: Training and Testing. During the training phase, the system learns to detect the category 
of violence present in a video by training classifiers with visual and audio features extracted from the training dataset. In the testing phase, the system is evaluated by calculating the accuracy of the system in detecting 
violence for a given video. Each of these phases is explained in detail in the following sections. Please refer to \fref{fig:overview} for the overview of the proposed approach. Finally, a section describing the metrics used 
for evaluating the system is presented.

\section{Training}
In this section, the details of the steps involved in the training phase are discussed. The proposed training approach has three main steps: Feature extraction, Feature Classification, and Feature fusion. Each of these three
steps is explained in detail in the following sections. In the first two steps of this phase, audio and visual features from the video segments containing violence and no violence are extracted and are used to train two-class
SVM classifiers. Then in the feature fusion step, feature weights are calculated for each violence type targeted by the system. These feature weights are obtained by performing a grid search on the possible combination of 
weights and finding the best combination which optimizes the performance of the system on the validation set. The optimization criteria here is the minimization of EER (Equal Error Rate) of the system. To find these weights,
a dataset disjoint from the training set is used, which contains violent videos of all targeted categories. Please refer to \cref{Chapter1} for details of targeted categories.

\subsection{Feature Extraction} \label{Feature Extraction}

Many researchers have tried to solve the Violence detection problem using different audio and visual features. A detailed information on violence detection related research is presented in \cref{Chapter2}.
In the previous works, the most common visual features used to detect violence are motion and blood and the most common audio feature used is the MFCC. Along with these three common low-level features, this proposed
approach also includes SentiBank (\citet{borth2013}), which is a visual feature representing sentiments in images. The details of each of the features and its importance in violence detection and the extraction
methods used are described in the following sections.

\subsubsection{MFCC-Features} \label{MFCC-Features}

Audio features play a very important role in detecting events such as gunshots, explosions etc, which are very common in violent scenes. Many researchers have used audio features for violence detection and have produced
good results. Even though some of the earlier works looked at energy entropy [\citet{nam1998}] in the audio signal, most of them used MFCC features to describe audio content in the videos. These MFCC features are commonly used
in voice and audio recognition. 

In this work, MFCC features provided in the VSD2014 dataset are used to train the SVM classifier while developing the system. During the evaluation, MFCC features are extracted from the audio stream of the input video, with  
window size set to the number of audio samples per frame in the audio stream. This is calculated by dividing the audio sampling rate with fps (frames per second) value of the video. For example, if the audio sampling rate is 
\num{44100} Hz and the video is encoded with \num{25} fps, then each window has \num{1764} audio samples. The window overlap region is set to zero and \num{22} MFCC are computed for each window. With this setup, a 
\num{22}-dimensional MFCC feature vector is obtained for each video frame. 

\subsubsection{Blood-Features} \label{Blood-Features}

Blood is the most common visible element in scenes with extreme violence. For example, scenes containing beating, stabbing, gunfire, and explosions. In many earlier works on violence detection, detection of pixels representing
blood is used as it is an important indicator of violence. To detect blood in a frame, a pre-defined color table is used in most of the earlier works, for example, \citet{nam1998} and \citet{lin2009}. Other approaches to 
detecting blood, such as the use of Kohonen's Self-Organizing Map (SOM)(\citet{clarin2005}), are also used in some of the earlier works. 

\begin{figure}[htbp]
\captionsetup{justification=justified,
singlelinecheck=false
}
\centering
\includegraphics[scale=0.5]{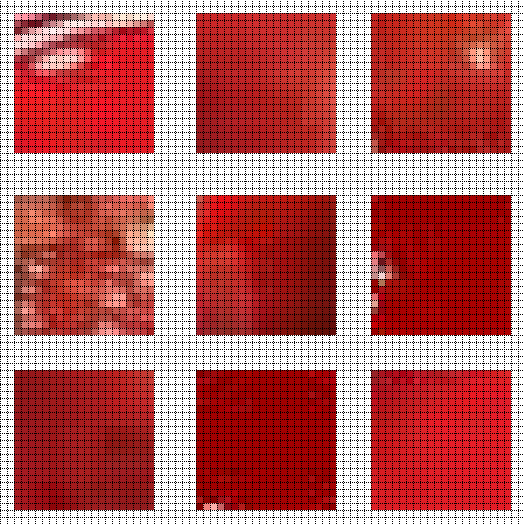}
\rule{\textwidth}{0.5pt}
\caption[Figure showing sample cropped regions of size \num{20} $\times$ \num{20} containing blood.]{Figure showing sample cropped regions of size \num{20} $\times$ \num{20} containing blood.}
\label{fig:bloodsample}
\end{figure}

In this work, a color model is used to detect pixels representing blood. It is represented using a three-dimensional histogram with one dimension each for red, green and blue values of the pixels. In each dimension, there are
\num{32} bins with each bin having width of \num{8} (\num{32} $\times$ \num{8} = \num{256}). This blood model is generated in two steps. In the first step, the blood model is bootstrapped by using the RGB (Red, Green, Blue) 
values of the pixels containing blood. The \num{3} dimensional binned histogram is populated with the RGB values of these pixels containing blood. The value in the bin to which a blood pixel belongs to is incremented by \num{1}
each time a new blood pixel is added to the model. Once a sufficient number of bloody pixels are used to fill the histogram, the values in the bins are normalized by the sum of all the values. The values in each of the bins
now represent the probability of a pixel showing blood given its RGB values. To fill the blood model, pixel containing blood are cropped from various images containing blood which are downloaded from Google. Cropping of the 
regions containing only blood pixels is done manually. Please refer to the image \fref{fig:bloodsample} for samples of the cropped regions, each of size \num{20} pixels $\times$ \num{20} pixels.

\begin{figure}[htbp]
\captionsetup{justification=justified,
singlelinecheck=false
}
\centering
\includegraphics[scale=0.35]{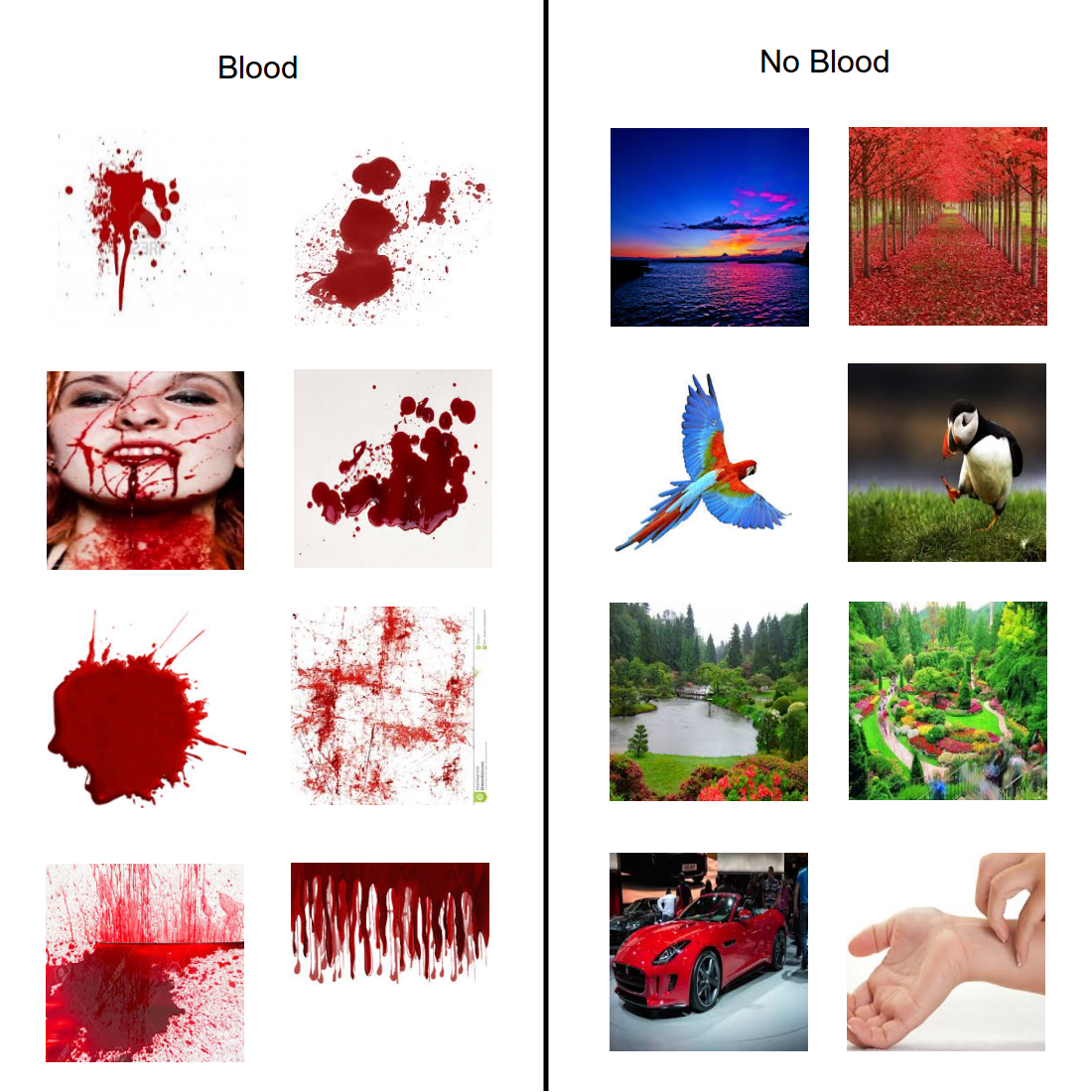}
\rule{\textwidth}{0.5pt}
\caption[Figure showing sample images downloaded from Google to generate blood and non-blood models.]{Figure showing sample images downloaded from Google to generate blood and non-blood models.}
\label{fig:googleimages}
\end{figure}

Once the model is bootstrapped, it is used to detect blood in the images downloaded from Google. Only pixels that have a high probability of representing blood are used to further extend the bootstrapped model. Downloading
the images and extending the blood model is done automatically. To download images from Google which contain blood, search words such as ``bloody images", ``bloody scenes", ``bleeding", ``real blood splatter",
``blood dripping" are used. Some of the samples of the downloaded images can be seen in the \fref{fig:googleimages}. Pixel values with high blood probability are added to the blood model until it has, at least, a million pixel
values. 

This blood model alone is not sufficient to accurately detect blood. Along with this blood model, there is a need for a non-blood model as well. To generate this, similar to the earlier approach, images are downloaded from 
Google which do not contain blood and the RGB pixel values from these images are used to build the non-blood model. Some samples images used to generate this non-blood model are shown in \fref{fig:googleimages}. Now using these
blood and non-blood models, the probability of a pixel representing blood is calculated as follows

\begin{equation}
    P(blood/pixel) = P_{blood}(pixel)/(P_{blood}(pixel) + P_{non-blood}(pixel))
\end{equation}

where P(blood/pixel) defines the probability of a pixel containing blood, P$_{blood}$(pixel) refers to the probability of blood for a given pixel and P$_{non-blood}$(pixel) corresponds to the probability of non-blood for a
pixel.

\begin{figure}[htbp]
\captionsetup{justification=justified,
singlelinecheck=false
}
\centering
\includegraphics[scale=0.2]{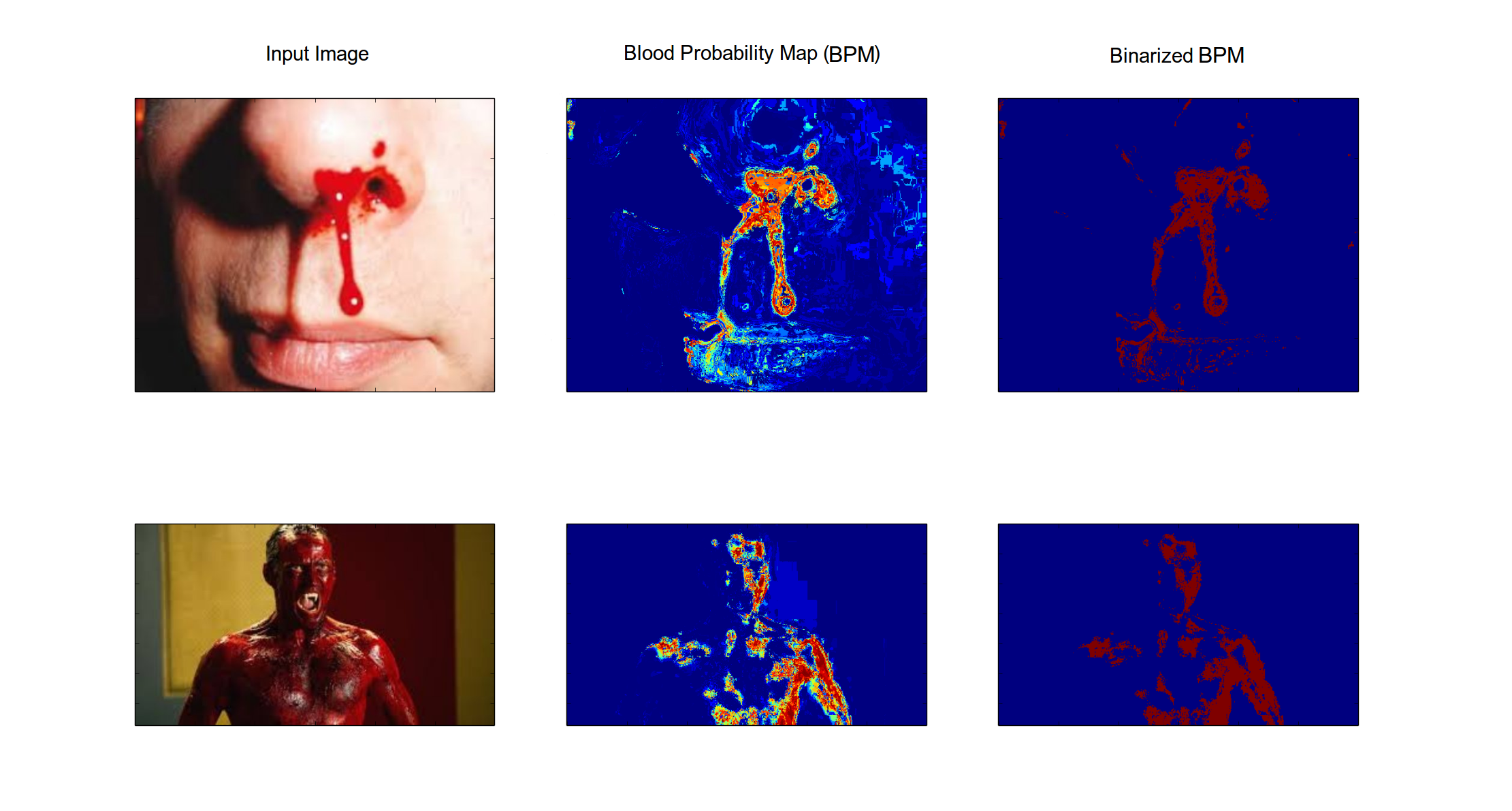}
\rule{\textwidth}{0.5pt}
\caption[Performance of Blood model in detecting blood.]{Figure showing the performance of the generated blood model in detecting blood. The first column has the input images, the second column has the blood probability maps
and the last column has the binarized blood probability maps. }
\label{fig:bloodmap}
\end{figure}

Using this formula, for a given image, the probability of each pixel representing blood is calculated and Blood Probability Map (BPM) is generated. This map has the same size as that of the input image and contains the 
blood probability values for every pixel. This BPM is binarized using a threshold value to generate the final binarized BPM. The threshold used to binarize the BPM is estimated (\citet{jones2002}). From this binarized BPM, 
a \num{1}-dimensional feature vector of length \num{14} is generated which contains the values such as the blood ratio, blood probability ratio, size of the biggest connected component, mean, variance etc. This feature vector
is extracted for each frame in the video and is used for training the SVM classifier. A sample image along with its BPM and binarized BPM are presented in \fref{fig:bloodmap}. It can be observed from this figure that this
approach has performed very well in detecting pixels containing blood.

\subsubsection{Motion-Features} \label{Motion-Features}

Motion is another widely used visual feature for violence detection. The work of \citet{deniz2014}, \citet{nievas2011} and \citet{hassner2012} are some of the examples in which motion is used as the main feature for 
violence detection. Here, motion refers to the amount of spatio-temporal variation between two consective frames in a video. Motion is considered a good indicator of violence as substantial amount of violence is expected 
in the scenes that contain violence. For example, in the scenes that contain person-on-person fights, there is fast movement of human body parts like legs and hands, and in scenes that contain explosions, there is a lot of
movement from the parts that are flying apart because of the explosion. 

The idea of using motion information for activity detection stems from psychology. Research on human perception has shown that the kinematic pattern of movement is sufficient for the perception of actions (\citet{blake2007}).
Research studies in computer vision (\citet{saerbeck2010}, \citet{clarke2005}, and \citet{hidaka2012}) have also shown that relatively simple dynamic features such as velocity and acceleration correlate to emotions perceived
by a human.

In this work, to calculate the amount of motion in a video segment, two different approaches are evaluated. The first approach is to use the motion information embedded inside the video codec and the next approach is to use 
optical flow to detect motion. These approahces are presented next.

\begin{figure}[htp]
\captionsetup{justification=justified,
singlelinecheck=false
}
\centering
\includegraphics[scale=0.5]{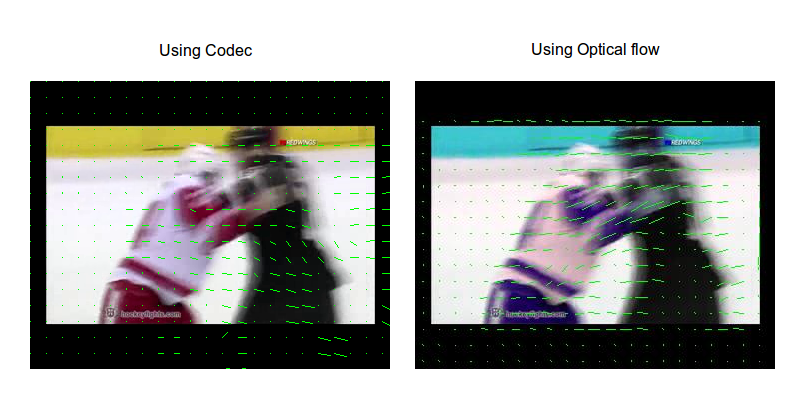}
\rule{\textwidth}{0.5pt}
\caption[Motion information from frames extracted using codec vs using optical flow.]{Motion information from frames extracted using codec vs using optical flow.}
\label{fig:motionmap}
\end{figure}

\paragraph{Using Codec} \label{Using Codec}
In this method, the motion information is extracted from the video codec. The magnitude of motion at each pixel per frame called the motion vector is retrieved from the codec. This motion vector is a two-dimensional vector
and has the same size as a frame from the video sequence. From this motion vector, a motion feature which represents the amount of motion in the frame is generated. To generate this motion feature, first the motion vector is 
divided into twelve sub-regions of equal sizes by slicing it along the x and y-axis into three and four regions respectively. The amount of motion along the x and y-axis at each pixel from each of these sub-regions are 
aggregated and these sums are used to generate a two-dimensional motion histogram for each frame. This histogram represents the motion vector for a frame. Refer to the image on the left in \fref{fig:motionmap} to see the
visualization of the aggregated motion vectors for a frame from a sample video. In this visualization, the motion vectors are aggregated for sub-regions of size \num{16} $\times$ \num{16} pixels. The magnitude and direction
of motion in these regions is represented using the length and orientation of the green dashed lines which are overlaid on the image.

\paragraph{Using Optical Flow} \label{Using Optical Flow}
The next approach to detect motion uses Optical flow (\citet{wiki:opticalflow}). Here, the motion at each pixel in a frame is calculated using Dense Optical Flow. For this, the implementation of Gunner Farneback's 
algorithm (\citet{farneback2003}) provided by OpenCV (\citet{opencv}) is used. The implementation is provided as a function in OpenCV and for more details about the function and the parameters, please refer to 
the documentation provided by OpenCV (\citet{opticalflowimpl}). The values \num{0.5}, \num{3}, \num{15}, \num{3}, \num{5}, \num{1.2} and \num{0} are passed to the function parameters pyr\_scale, levels, win-size, iterations,
poly\_n, poly\_sigma and flags respectively. Once the motion vectors at every pixel are calculated using Optical flow, the motion feature from a frame is extracted using the same process mentioned in the above
\sref{Using Codec}. Refer to the image on the rights in \fref{fig:motionmap} to get an impression of the aggregated motion vectors extracted from a frame. The motion vectors are aggregated for sub-regions of
size $\num{16} \times \num{16}$ pixels as in the previous approach to provide a better comparison between the features extracted by using Codec information and Optical flow.

After the evaluation of both these approaches to extract motion information from videos, the following observations are made. First, extracting motion from Codecs is much faster than using optical flow as the motion vectors are
precalculated and stored in video codecs. Second, motion extraction using optical flow is not very efficient when there are blurred regions in a frame. This blur is usually caused by sudden motions in a scene, which is very
common in scenes containing violence. Hence, the use of optical flow for extracting motion information to detect violence is not a promising approach. Therefore, in this work information stored in the video codecs is used to
extract motion features. The motion features are extracted from each frame in the video and are used to train an SVM classifier.

\subsubsection{SentiBank-Features} \label{SentiBank-Features}

In addition to the aforementioned low-level features, the SentiBank feature introduced by \citet{borth2013} is also applied. SentiBank is a mid-level representation of visual content based on the large-scale Visual 
Sentiment Ontology (VSO) \footnote{http://visual-sentiment-ontology.appspot.com}. SentiBank consists of \num{1200} semantic concepts and corresponding automatic classifiers, each being defined as an Adjective Noun Pair (ANP).
Such ANPs combine strong emotional adjectives to be linked with nouns, which correspond to objects or scenes (e.g. “beautiful sky”, “disgusting bug”, or “cute baby”). Further, each ANP (1) reflects a strong sentiment, 
(2) has a link to an emotion, (3) is frequently used on platforms such as Flickr or YouTube and (4) has a reasonable detection accuracy. Additionally, the VSO is intended to be comprehensive and diverse enough to cover 
a broad range of different concept classes such as people, animals, objects, natural or man-made places and, therefore, provides additional insights about the type of content being analyzed. Because SentiBank demonstrated
its superior performance as compared to low-level visual features on the analysis of sentiment \citet{borth2013}, it is used now for the first time to detect complex emotion such as violence from video frames. 

SentiBank consists of \num{1200} SVMs, each trained to detect one of the \num{1200} semantic concepts from an image. Each SVM is a binary classifier which gives a binary output 0/1 depending on whether or not the
image contains a specific sentiment. For a given frame in a video, a vector containing the output of all \num{1200} SVMs is considered as the SentiBank feature. To extract this feature, a python-based implementation is
utilized. For training the SVM classifier, the SentiBank features extracted from each frame in the training videos are used. The SentiBank feature extraction takes few seconds as it involves collecting output from \num{1200}
pre-trained SVMs. To reduce the time taken for feature extraction, the SentiBank feature for each of the frame is extracted in parallel using multiprocessing.

\subsection{Feature Classification} \label{Feature Classification}

The next step in the pipeline after feature extraction is feature classification and this section provides the details of this step. The selection of classifier and the training techniques used play a very important role in 
getting good classification results. In this work, SVMs are used for classification. The main reason behind this choice is the fact that the earlier works on violence detection have used SVMs to classify audio and visual 
features and have produced good results. In almost all the works mentioned in \cref{Chapter2} SVMs are used for classification, even though they may differ in the kernel functions used. 

From all the videos available in the training set, audio and visual features are extracted using the process described in the \sref{Feature Extraction}. These features are then divided into two sets, one to train the
classifier and the other to test the classification accuracy of the trained classifier. As the classifiers used here are SVMs, a choice has to be made about what kernel to use and what kernel parameters to set. To find the 
best kernel type and kernel parameters, a grid search technique is used. In this grid search, Linear, RBF (Radial Basis Function), and Chi-Square kernels along with a range of value for their parameters are tested, to find 
the best combination which gives the best classification results. Using this approach, four different classifiers are trained, one for each feature type. These trained classifiers are then used in finding the feature weights 
in the next step. In this work, the SVM implementation provided by scikit-learn (\citet{scikit-learn}) and LibSVM (\citet{Chang2011}) are used.

\subsection{Feature Fusion} \label{Feature Fusion}

In the feature fusion step, the output probabilities from each of the feature classifiers are fused to get the final score of the violence in a video segment along with the class of violence present in it. This fusion
is done by calculating the weighted sum of the probabilities from each of the feature classifiers. To detect the class of violence to which a video belongs, the procedure is as follows. First, the audio and visual features are
extracted from the videos belonging to each of the targeted violence classes. These features are then passed to the trained binary SVM classifiers to get the probabilities of each of the video containing violence. Now, these
output probabilities from each of the feature classifiers are fused by assigning each feature classifier a weight for each class of violence and calculating the weighted sum. The weights assigned to each of the feature 
classifiers represent the importance of a feature in detecting a specific class of violence. These feature weights have to be adjusted appropriately for each violence class for the system to detect the correct class of 
violence. 

There are two approaches to finding the weights. The first approach is to manually adjust weights of a feature classifier for each violence type. This approach needs a lot of intuition about the importance of a feature in 
detecting a class of violence and is very error prone. The other approach is to find the weights using a grid-search mechanism where a set of weights is sampled from the range of possible weights. In this case, the range of
possible weights for each feature classifier is [\num{0},\num{1}], subjected to the constraint that the sum of weights of all the feature classifiers amounts to \num{1}. In this work, the latter approach is used and all the
weight combinations which amount to \num{1} are enumerated. Each of these weight combinations is used to calculate the weighted sum of classifier probabilities for a class of violence and the weights from the weight 
combination which produces the highest sum is assigned to each of the classifiers for the corresponding class of violence. To calculate these weights, a dataset which is different from the training set is used, in order to 
avoid over-fitting of weights to the training set. The dataset used for weight calculation has videos from all the classes of violence targeted in this work. It is important to note that, even though each of the trained SVM 
classifiers are binary in nature, the output values from these classifiers can be combined using weighted sum to find the specific class of violence to which a video belongs.   

\section{Testing} \label{Testing}

In this stage, for a given input video, each segment containing violence is detected along with the class of violence present in it. For a given video, the following approach is used to detect the segments that contain 
violence and the category of violence in it. First, the visual and audio features are extracted from one frame every \num{1}-second starting from the first frame of the video, rather than extracting features from every frame.
These frames from which the features are extracted, represent a \num{1}-second segment of the video. The features from these 1-second video segments are then passed to the trained binary SVM classifiers to get the scores for 
each video segment to be violent or non-violent. Then, weighted sums of the output values from the individual classifiers are calculated for each violence category using the corresponding weights found during fusion step.
Hence, for a given video of length `X' seconds, the system outputs a vector of length `X'. Each element in this vector is a dictionary which maps each violence class with a score value. The reason for using this approach
is two fold, first to detect time intervals in which there is violence in the video and to increase the speed of the system in detecting violence. The feature extraction, especially extracting the Sentibank feature, is time-
consuming and doing it for every frame will make the system slow. But this approach has a negative effect on the accuracy of the system as it detects violence not for every frame but for every second. 

\section{Evaluation Metrics} \label{Evaluation Metrics}

There are many metrics that can be used to measure the performance of a classification systems. Some of the measures used for binary classification are Accuracy, Precision, Recall (Sensitivity), Specificity, F-score, Equal Error
Rate (EER), and Area Under the Curve (AUC). Some other measures such as Average Precision (AP) and Mean Average Precision (MAP) are used for systems which return a ranked list as a result to a query. Most of these measures that
are increasingly used in Machine Learning and Data Mining research are borrowed from other disciplines such as Information Retrieval (\citet{Rijsbergen1979}) and Biometrics. For a detailed discussion on these measures, 
refer to the works of \citet{parker2011} and \citet{sokolova2009}. The ROC (Receiver Operating Characteristic) curve is another widely used method for evaluating or comparing binary classification systems. Measures such as
AUC and EER can be calculated from the ROC curve.

In this work, ROC curves are used to: (i) Compare the performance of individual classifiers. (ii) Compare the performance of the system in detecting different classes of violence in the Multi-Class classification task. 
(iii) Compare the performance of the system on Youtube and Hollywood-Test dataset in the Binary classification task. Other metrics that are used here are, Precision, Recall and EER. These measures are used as these are the
most commonly used measures in the previous works on violence detection.
In this system, the parameters (fusion weights) are adjusted to minimize the EER. 

\section{Summary}

In this chapter, a detailed description of the approach followed in this work to detect violence is presented. The first section deals with the training phase and the second section deals with the testing phase. In the first 
section, different steps involved in the training phase are explained in detail. First the extraction of audio and visual features is discussed and the details of what features are used and how they are extracted are presented.
Next, the classification techniques used to classify the extracted features are discussed. Finally, the process used to calculate feature weights for feature fusion is discussed. In the second section, the process used during
the testing phase to extract video segments containing violence and to detect the class of violence in these segments is discussed.

To summarize, the steps followed in this approach are feature extraction, feature classification, feature fusion and testing. The first three steps constitute the training phase and the final step is the testing phase. In the 
training phase, audio and visual features are extracted from the video and they are used to train binary SVM classifiers one for each feature. Then, a separate dataset is used to find the feature weights which minimize the EER
of the system on the validataion dataset. In the final testing phase, first the visual and audio features are extracted one per a 1-second video segment of the input test video. Then, these features are passed 
to the trained SVM classifiers to get the probabilities of these features representing violence. A weighted sum of these output probabilities is calculated for each violence type using the weights obtained in the feature
fusion step. The violence type for which the weighted sum is maximum is assigned as a label to the corresponding 1-second video segment. Using these labels the segments containing violence and the class of violence contained 
in them is presented as an output by the system. The experimental setup and evaluation of this system are presented in the next chapter.

\chapter{Experiments and Results} 

\label{Chapter4} 

\lhead{Chapter 4. \emph{Experiments and Results} }

In this chapter, details of the experiments conducted to evaluate the performance of the system in detecting violent content in videos are presented. The first section deals with the datasets used for this work, the next
section describes the experimental setup and finally in the last section, results of the experiments performed are presented.

\section{Datasets} \label{Datasets}

In this work, data from more than one source has been used to extract audio and visual features, train the classifiers and to test the performance of the system. The two main datasets used here are the Violent Scene Dataset
(VSD) and the Hockey Fights dataset. Apart from these two datasets, images from websites such as Google Images\footnote{http://www.images.google.com\label{google_images_note}} are also used. Each of these datasets and their 
use in this work is described in detail in the following sections.

\subsection{Violent Scene Dataset} \label{VSD2014}

Violent Scene Dataset (VSD) is an annotated dataset for violent scene detection in Hollywood movies and videos from the web. It is a publicly available dataset specifically designed for the development of content-based 
detection techniques targeting physical violence in movies and videos from the websites such as YouTube\footnote{http://www.youtube.com}. The VSD dataset was initially introduced by \citet{demarty2011} in the framework of the 
MediaEval benchmark initiative, which serves as a validation framework for the dataset and establishes a state of the art baseline for the violence detection task. The latest version of the dataset VSD2014 is a considerable
extension of its previous versions (\citet{demarty2014} , \citet{demarty2014multimodal} and \citet{demarty2014benchmarking}) in several regards.  First, to annotate the movies and user-generated videos, violence definition 
which is closer to the targeted real-world scenario is used by focusing on physical violence one would not let a \num{8}-year-old child watch. Second, the dataset has a substantial set of \num{31} Hollywood movies. 
Third, VSD2014 includes \num{86} web video clips and their meta-data retrieved from YouTube to serve for testing the generalization capabilities of the system developed to detect violence. Fourth, it includes state-of-the-art
audio-visual content descriptors. The dataset provides annotations of violent scenes and of violence-related concepts for a collection of (i) Hollywood movies and (ii) user-generated videos shared on the web. In addition to 
the annotations, pre-computed audio and visual features and various meta-data are provided.

\begin{table}
\captionsetup{justification=justified,
singlelinecheck=false
} 
 \begin{center} 
 \caption[Statistics of the movies and videos in the VSD2014 subsets.]{ Statistics of the movies and videos in the VSD2014 subsets. All values are given in Seconds.}
 \label{table:vsd}
 \begin{tabular}{ p{0.30\linewidth}  >{\centering\arraybackslash} p{0.12\linewidth}  >{\centering\arraybackslash} p{0.30\linewidth}  >{\centering\arraybackslash} p{0.12\linewidth}  } 
    \hline
    \textbf{Name} & \textbf{Duration} & \textbf{Fraction of Violence} (\%) & \textbf{Avg. Duration}\\
    \hline
    \multicolumn{4}{c}{ \textbf{Hollywood: Development}}\\
    \hline
    Armageddon & 8,680.16 & 7.78 & 25.01 \\
    Billy Elliot & 6,349.44 & 2.46 & 8.68 \\
    Dead Poets Society & 7,413.20 & 0.58 & 14.44 \\
    Eragon & 5,985.44 & 13.26 & 39.69 \\
    Fantastic Four 1 & 6,093.96 & 20.53 & 62.57 \\
    Fargo & 5,646.40 & 15.04 & 65.32 \\
    Fight Club & 8,004.50 & 15.83 & 32.51 \\
    Forrest Gump & 8,176.72 & 8.29 & 75.33 \\
    Harry Potter 5 & 7,953.52 & 5.44 & 17.30 \\
    I am Legend & 5,779.92 & 15.64 &  75.36 \\
    Independence Day & 8,833.90 & 13.13 & 68.23 \\
    Legally Blond & 5,523.44 & 0.00 & 0.00 \\
    Leon & 6,344.56 & 16.36 & 41.52 \\
    Midnight Express & 6,961.04 & 7.12 & 24.80 \\
    Pirates of the Caribbean & 8,239.40 & 18.15 & 49.85 \\
    Pulp Fiction & 8,887.00 & 25.05 & 202.43 \\
    Reservoir Dogs & 5,712.96 & 30.41 & 115.82 \\
    Saving Private Ryan & 9,751.00 & 33.95 & 367.92 \\
    The Bourne Identity & 6,816.00 & 7.18 & 27.21 \\
    The God Father & 10,194.70 & 5.73 & 44.99 \\
    The Pianist & 8,567.04 & 15.44 & 69.64 \\
    The Sixth Sense & 6,178.04 & 2.00 & 12.40 \\
    The Wicker Man & 5,870.44 & 6.44 & 31.55 \\
    The Wizard of Oz & 5,859.20 & 1.02 & 8.56 \\
    \hline
    \textbf{Total} & 180,192.40 & 12.35 \\
    \\
    \hline
    \multicolumn{4}{c}{ \textbf{Hollywood: Test}}\\
    \hline
    8 Mile & 6,355.60 & 4.70 & 37.40 \\
    Braveheart & 10,223.92 & 21.45 & 51.01 \\
    Desperado & 6,012.96 & 31.94 & 113.00 \\
    Ghost in the Shell &  4966.00 & 9.85 & 44.47 \\
    Jumanji &  5993.96 & 6.75 & 28.90 \\
    Terminator 2 &  8831.40 & 24.89 & 53.62 \\
    V for Vendetta & 7625.88 & 14.27 & 25.91 \\
    \hline
    \textbf{Total} & 50,009.72 & 17.18 \\
    \\
    \hline
    \multicolumn{4}{c}{ \textbf{YouTube: Generalization}}\\
    \hline
    Average & 109.76 & 31.69 & 26.62 \\
    Std.dev & 68.05 & 36.28 & 50.41 \\
    \hline
    \textbf{Total} & 9,439.39 & 31.69
    
 \end{tabular}
 \end{center}
 
\end{table}

The VSD2014 dataset is split into three different sub-sets, called Hollywood: Development, Hollywood: Test, and YouTube: Generalization. Please refer to \tref{table:vsd} for an overview of the three subsets and basic 
statistics, including duration, the fraction of violent scenes (as percentage on a per-frame-basis), and the average length of a violent scene. The content of the VSD2014 dataset is categorized into three types: movies/videos,
features, and annotations.

The Hollywood movies included in the dataset are chosen such that they are from different genres and have diversity in the types of violence they contain. Movies ranging from extremely violent to virtually no violent content
are selected to create this dataset. The selected movies also contain a wide range of violence types. For example, war movies, such as Saving Private Ryan, contain specific gunfights and battle scenes involving lots of people,
with a loud and dense audio stream containing numerous special effects. Action movies, such as the Bourne Identity, contain scenes of fights involving only a few participants, possibly hand to hand. Disaster movies, such as 
Armageddon, show the destruction of entire cities and contain huge explosions. Along with these, a few completely nonviolent movies are also added to the dataset to study the behavior of algorithms on such content. As the 
actual movies can not be provided in the dataset due to copyright issues, annotations for \num{31} movies, \num{24} in the Hollywood: Development and \num{7} in the Hollywood: Test set are provided. The YouTube: Generalization
set contains video clips shared on YouTube under Creative Commons license. A total of \num{86} clips in MP4 format is included in the dataset. Along with the video meta-data such as video identifier, publishing date, category,
title, author, aspect ratio, duration etc., are provided as XML files.

In this dataset, a common set of audio and visual descriptors are provided. Audio features such as amplitude envelop (AE), root-mean-square energy (RMS), zero-crossing rate (ZCR), band energy ratio (BER), spectral centroid (SC), 
frequency bandwidth (BW), spectral flux (SF), and Mel-frequency cepstral coefficients (MFCC) are provided on a per-video-frame-basis. As audio has a sampling rate of \num{44100} Hz and the videos are encoded with \num{25} fps,
a window of size \num{1764} audio samples in length is considered to compute these features and \num{22} MFCCs are computed for each window while all other features are \num{1}-dimensional. Video features provided in the 
dataset include color naming histograms (CNH), color moments (CM), local binary patterns (LBP), and histograms of oriented gradients (HOG). Audio and visual features are provided in Matlab version 7.3 MAT files, which 
correspond to HDF5 format. 

The VSD2014 dataset contains binary annotations of all violent scenes, where a scene is identified by its start and end frames. These annotations for Hollywood movies and YouTube videos are created by several human assessors 
and are subsequently reviewed and merged to ensure a certain level of consistency. Each annotated violent segment contains only one action, whenever this is possible. In cases where different actions are overlapping, the 
segments are merged. This is indicated in the annotation files by adding the tag “multiple action scene”. In addition to binary annotations of segments containing physical violence, annotations also include high-level concepts
for \num{17} movies in the Hollywood: Development set. In particular, \num{7} visual concepts and \num{3} audio concepts are annotated, employing a similar annotation protocol as used for violent/non-violent annotations. 
The concepts are the presence of blood, fights, presence of fire, presence of guns, presence of cold arms, car chases, and gory scenes, for the visual modality; the presence of gunshots, explosions, and screams for the audio
modality.

A more detailed description of this dataset is provided by \citet{schedl6vsd2014} and for the details about each of the violence classes, please refer to \citet{demarty2014}.

\begin{figure}[htbp]
\captionsetup{justification=justified,
singlelinecheck=false
}
\centering
\includegraphics[scale=0.25]{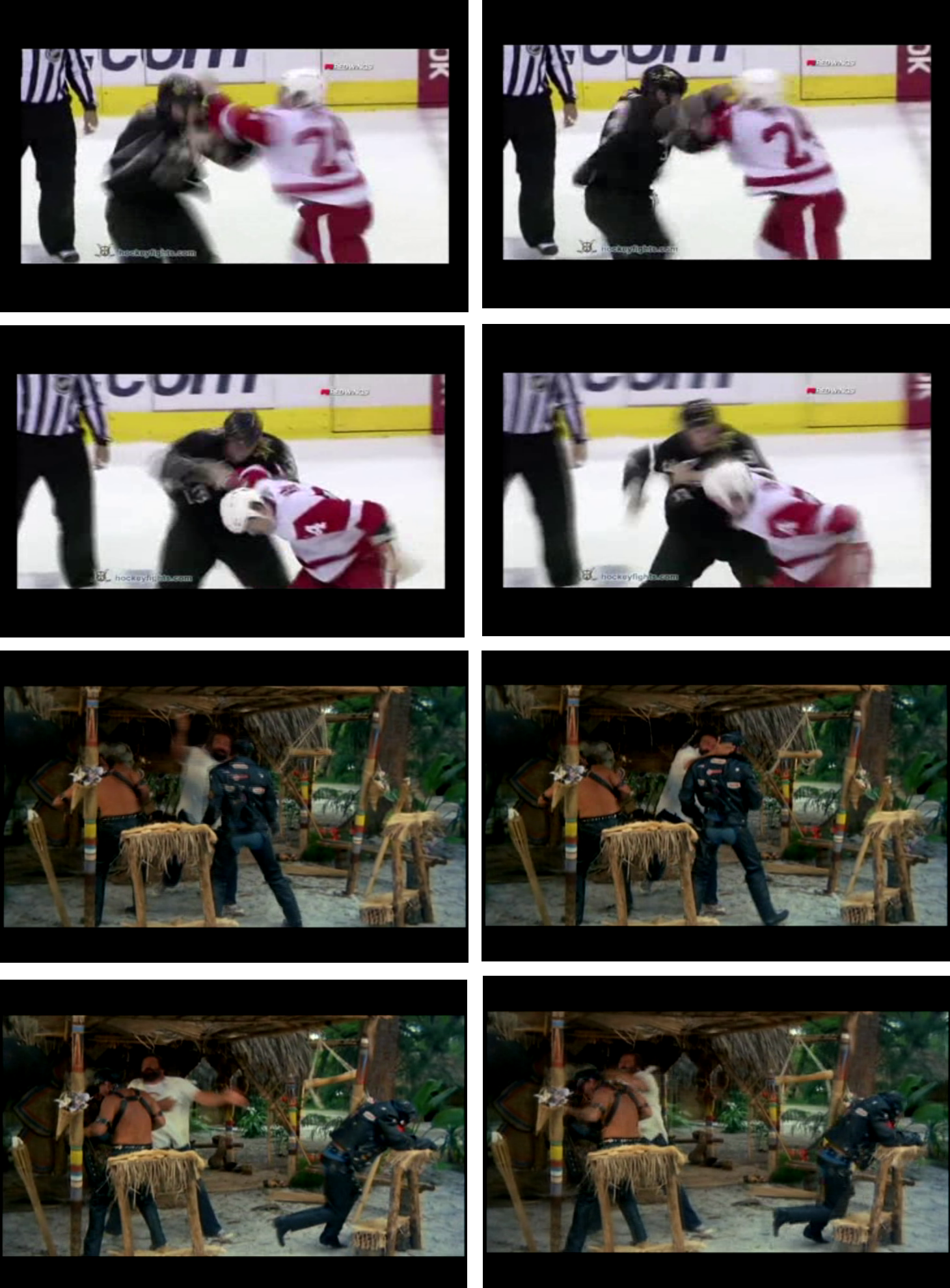}
\rule{\textwidth}{0.5pt}
\caption[Sample frames from the fight videos in the Hockey (top) and action movie (bottom) datasets.]{Sample frames from the fight videos in the Hockey (top) and action movie (bottom) datasets.}
\label{fig:fights}
\end{figure}

\subsection{Fights Dataset}

This dataset is introduced by \citet{nievas2011} and it is created specifically for evaluating fight detection systems. This dataset consists of two parts, the first part (“Hockey”) consists of \num{1000} clips at a resolution
of \num{720} $\times$ \num{576} pixels, divided into two groups, \num{500} fights, and \num{500} non-fights, extracted from hockey games of the National Hockey League (NHL). Each clip is limited to \num{50} frames and
resolution lowered to \num{320} $\times$ \num{240}. The second part (“Movies”) consists of \num{200} video clips, \num{100} fights, and \num{100} non-fights, in which fights are extracted from action movies and the non-fight
videos are extracted from public action recognition datasets. Unlike the hockey dataset, which was relatively uniform both in format and content, these videos depict a wider variety of scenes and were captured at different
resolutions. Refer to \fref{fig:fights} for some of the frames showing fights from the videos in the two datasets. This dataset is available on-line for
download\footnote{http://visilab.etsii.uclm.es/personas/oscar/FightDetection/index.html}.

\subsection{Data from Web}

Images from Google are used in developing the color models (\sref{Blood-Features}) for the classes blood and non-blood, which are used in extracting blood feature descriptor for each frame in a video. The images containing 
blood are downloaded from \mbox{Google Images \footnotemark[1]\footnotetext[1]{http://www.images.google.com}} using query words such as ``bloody images", ``bloody scenes", ``bleeding", ``real blood splatter" etc. Similarly, 
images containing no blood are downloaded using search words such as ``nature",``spring",``skin",``cars" etc.

The utility to download images from Google, given a search word, was developed in Python using the library Beautiful Soup (\citet{richardson2013}). For each query, the response contained about \num{100} images of which
only the first \num{50} were selected for download and saved in a local file directory. Around \num{1000} images were downloaded in total, combining both blood and non-blood classes. The average dimensions of the images 
downloaded are $\num{260}\times\num{193}$ pixels with a file size of around 10 Kilobytes. Refer to \fref{fig:googleimages} for some of the sample images used in this work.

\section{Setup} \label{Setup}

In this section, details of the experimental setup and the approaches used to evaluate the performance of the system are presented. In the following paragraph, partitioning of the dataset is discussed and the later paragraphs
explain the evaluation techniques.

As mentioned in the earlier \sref{Datasets}, data from multiple sources is used in this system. The most important source is the VSD2014 dataset. It is the only publicly available dataset which provides 
annotated video data with various categories of violence and it is the main reason for using this dataset in developing this system. As explained in the previous \sref{VSD2014}, this dataset contains three subsets, 
Hollywood: Development, Hollywood: Test and YouTube: Generalization. In this work all the three subsets are used. The Hollywood: Development subset is the only dataset which is annotated with different violence classes. This
subset consisting of \num{24} Hollywood movies is partitioned into \num{3} parts. The first part consisting of \num{12} movies (Eragon, Fantastic Four 1, Fargo, Fight Club, Harry Potter 5, I Am Legend, Independence Day, 
Legally Blond, Leon, Midnight Express, Pirates Of The Caribbean, Reservoir Dogs) is used for training the classifiers. The second part consisting of \num{7} movies (Saving Private Ryan, The Bourne Identity, The God Father,
The Pianist, The Sixth Sense, The Wicker Man, The Wizard of Oz) is used for testing the trained classifiers and to calculate weights for each violence type. The final part consisting of \num{3} movies (Armageddon, Billy Elliot,
and Dead Poets Society) is used for evaluation. The Hollywood: Test and the YouTube: Generalization subsets are also used for evaluation, but for a different task. The following paragraphs provide details of the evaluation 
approaches used.

To evaluate the performance of the system, two different classification tasks are defined. In the first task, the system has to detect specific category of violence present in a video segment. The second task is more 
generic where the system has to only detect the presence of violence. For both these tasks, different datasets are used for evaluation. In the first task which is a multi-class classification task, the validation set 
consisting of \num{3} Hollywood movies (Armageddon, Billy Elliot, and Dead Poets Society) is used. In this subset, each frame interval containing violence is annotated with the class of violence that is present. Hence, this 
dataset is used for this task. These \num{3} movies were neither used for training, testing of classifiers nor for weight calculation so that the system can be evaluated on a purely new data. The procedure illustrated in
\fref{fig:overview} is used for calculating the probability of a video segment to belong to a specific class of violence. The output probabilities from the system and the ground truth information are used to generate ROC 
(Receiver Operating Characteristic) curves and to assess the performance of the system.

In the second task, which is a binary classification task, Hollywood: Test and the YouTube: Generalization subsets of the VSD2104 dataset are used. The Hollywood: Test subset consists of \num{8} Hollywood movies and  the 
YouTube: Generalization subset consists for \num{86} videos from YouTube. In both these subsets the frame intervals containing violence are provided as annotations and no information about the class of violence is provided. 
Hence, these subsets are used for this task. In this task, similar to the previous one, the procedure illustrated in \fref{fig:overview} is used for calculating the probability of a video segment to belong to a specific class
of violence. For each video segment, the maximum probability obtained for any of the violence class is considered to be the probability of it being violent. Similar to above task, ROC curves are generated from these 
probability values and the ground truth from the dataset.

In both these tasks, first all the features are extracted from the training and testing datasets. Next, the training and testing datasets are randomly sampled to get an equal amount of positive and negative samples. \num{2000}
feature samples are selected for training and \num{3000} are selected for testing. As mentioned above, disjoint training and testing sets are used to avoid testing on training data. In both the tasks, SVM classifiers with 
Linear, Radial Basis Function and Chi-Square kernels are trained for each feature type and the classifiers with good classification scores on the test set are selected for the fusion step. In the fusion step, the weights for
each violence type are calculated by grid-searching the possible combinations which maximize the performance of the classifier. The EER (Equal Error Rate) measure is used as the performance measure.

\section{Experiments and Results} \label{Experiments and Results} 

In this section, the experiments and their results are presented. First, the results of the multi-class classification task are presented, followed by the results of the binary classification task.  

\subsection{Multi-Class Classification} \label{MultiClass Classification}

In this task, the system has to detect the category of violence present in a video. The violence categories targeted in this system are Blood, Cold arms, Explosions, Fights, Fire, Firearms, Gunshots, Screams. As
mentioned in the \cref{Chapter1}, these are the subset of categories of violence that are defined in the VSD2014. Apart from these eight categories, Car Chase, and Subjective Violence are also defined in VSD2014, which are not used
in this work as there were not enough video segments tagged with these categories in the dataset. This task is very difficult as detection of sub-categories of violence adds more complexity to the complicated problem of 
violence detection. The attempt to detect fine-grained concepts of violence by this system is novel and there is no existing system which does this task.

As mentioned in \cref{Chapter3}, this system uses weighted decision fusion approach to detect multiple classes of violence where weights for each violence category are learned using a grid-search technique. Please refer
to \sref{Feature Fusion} for more details about this approach. In \tref{table:weights}, the weights for each violence class which is found using this grid-search technique are presented.

\begin{table}
\captionsetup{justification=justified,
singlelinecheck=false
}
\begin{center} 
 \caption[Classifier weights obtained for each violence class using Grid-Search Technique] { Classifier weights obtained for each of the violence class using Grid-Search technique. Here the criteria for selecting the weights
 for a violence class was to find the weights which minimize the EER for that violence class. }
 \label{table:weights}
 \begin{tabular}{  l  c  c  c c} 
  
    \hline
    \textbf{Violence Class} & \textbf{Audio} & \textbf{Blood} & \textbf{Motion} & \textbf{SentiBank}\\
    \hline
    GunShots & 0.50 & 0.45 & 0.00 & 0.05 \\
    Fights & 0.40 & 0.05 & 0.25 & 0.30 \\
    Explosions & 0.90 & 0.00 & 0.00 & 0.10\\
    Fire & 0.05 & 0.05 & 0.05 & 0.85 \\
    Cold arms & 0.05 & 0.00 & 0.00 & 0.95\\
    Firearms & 0.05 & 0.30 &  0.05 & 0.60 \\
    Blood & 0.00 & 0.05 & 0.00 & 0.95\\
    Screams & 0.05 & 0.20 & 0.00 & 0.75 \\
    \hline
    
 \end{tabular}
 \end{center}
 
\end{table}

\begin{figure}[htbp]
\captionsetup{justification=justified,
singlelinecheck=false
}
\centering
\includegraphics[scale=1.0]{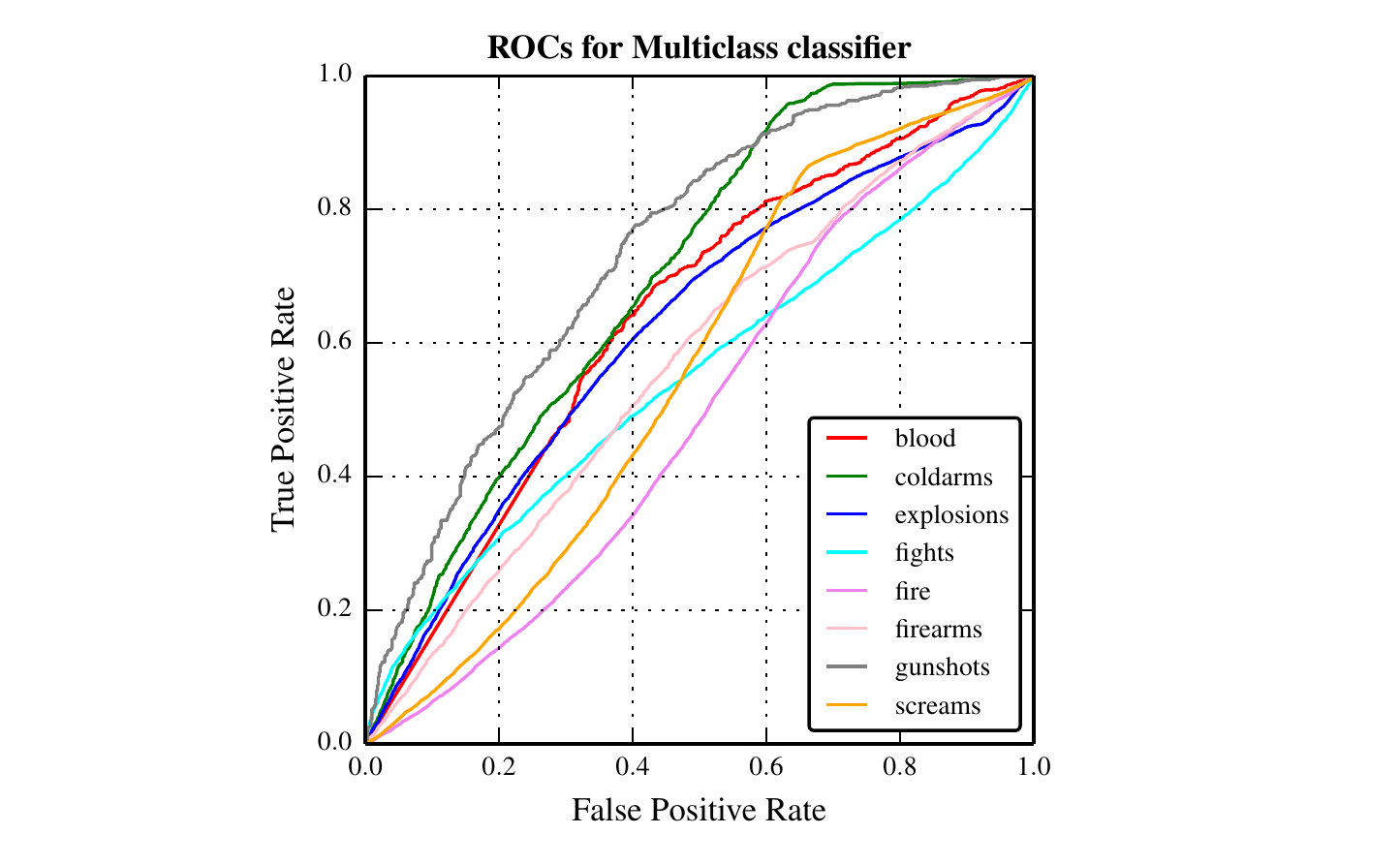}
\rule{\textwidth}{0.5pt}
\caption[Performance of the system in the Multi-Class Classification task.]{Performance of the system in the Multi-Class Classification task.}
\label{fig:multiclass}
\end{figure}

These weights are used to get the weighted sum of output values of binary feature classifiers for each violence category. The category with the highest sum is then the category of violence present in that video segment.
If the output sum is less than 0.5 then the video segement is categorised as Non-Violent. The video segments in the validation set are classified using this approach and the results are presented in the  \fref{fig:multiclass}.
In the figure, each curve represents the ROC curve for each of the violence categories.

\subsection{Binary Classification} \label{Binary Classification}

\begin{figure}[htbp]
\captionsetup{justification=justified,
singlelinecheck=false
}
\centering
\includegraphics[scale=1.0]{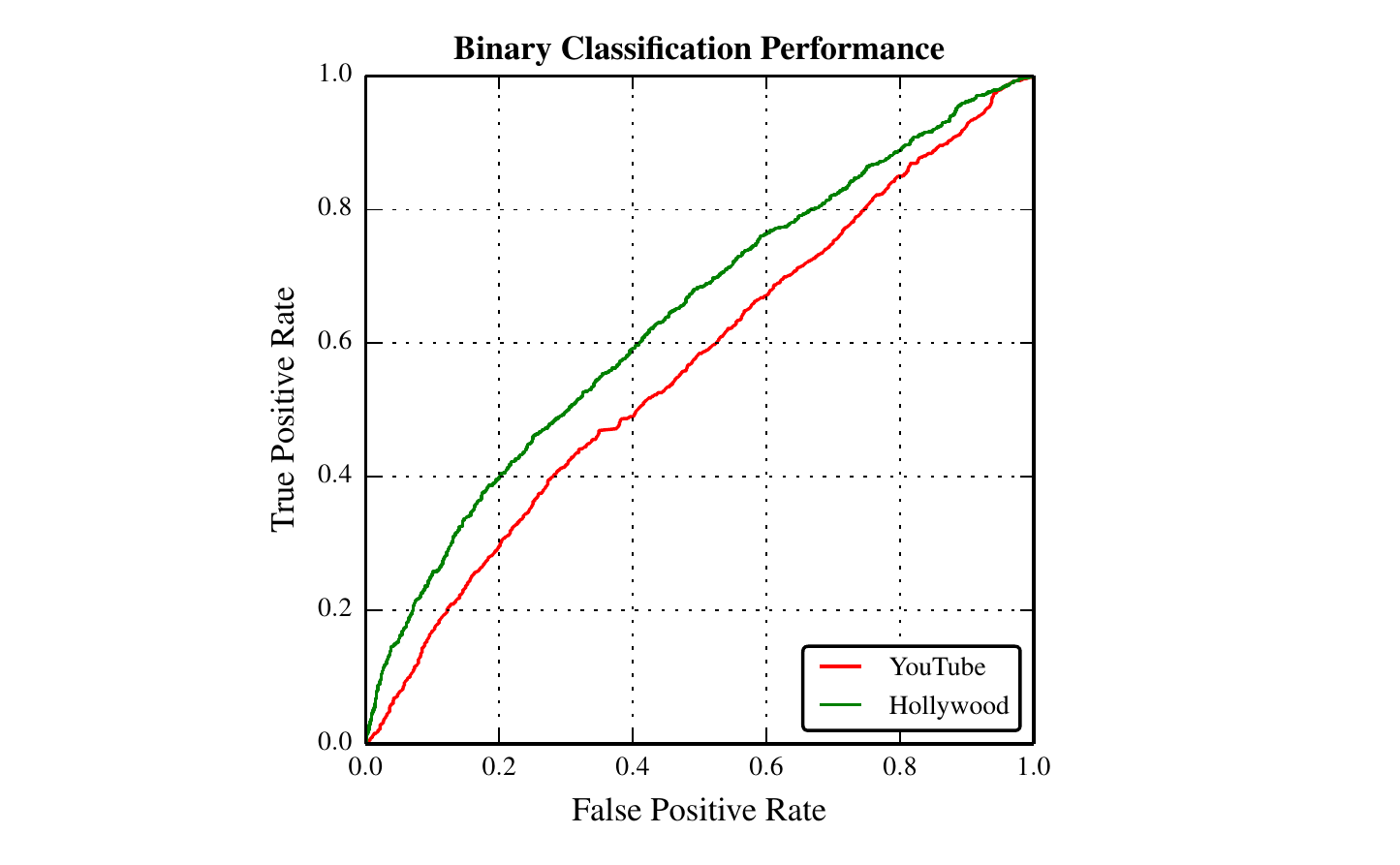}
\rule{\textwidth}{0.5pt}
\caption[Performance of the system in the Binary Classification task.]{Performance of the system in the Binary Classification task.}
\label{fig:binary}
\end{figure}

In this binary classification task, the system is expected to detect the presence of violence without having to find the category. Similar to the previous task, the output probabilities of binary feature classifiers are 
combined using a weighted sum approach and the output probabilities of the video segment to belong to each of the violence classes are calculated. If the maximum probability for any of the class exceeds \num{0.5} then the
video segment is categorized as violence or else it is categorized as non-violence. As mentioned in \sref{Setup}, this task is performed on YouTube-Generalization and Hollywood-Test datasets. The \fref{fig:binary} provides the
results of this task on both the datasets. Two ROC curves one for each of the datasets are used to represent the performance of the system. Using \num{0.5} as the threshold to make the decision of whether the video segment 
contains violence or not, the precision, recall and accuracy values are calculated. Please refer to \tref{table:results} for the obtained results.

\begin{table}
\captionsetup{justification=justified,
singlelinecheck=false
}
 \begin{center} 
 \caption[Classification results obtained using the proposed approach.]{Classification results obtained using the proposed approach.}
 \label{table:results}
 \begin{tabular}{  l  c  c  c c} 
  
    \hline
    \textbf{Test Set} & \textbf{Precision} & \textbf{Recall} & \textbf{Accuracy} & \textbf{EER} \\
    \hline
    YouTube-Gen & 51.4\% &  61.3\% & 55.6\% & 45.0\% \\
    Hollywood-Test & 62.7\% & 85.1\% & 59.2\% & 44.0\% \\
    \hline
    
 \end{tabular}
 \end{center}
 
\end{table}

\begin{table}
\captionsetup{justification=justified,
singlelinecheck=false
}
 \begin{center} 
 \caption[Classification results obtained by the best performing teams from MediaEval-2014.]{Classification results obtained by the best performing teams from MediaEval-2014 (\citet{schedl6vsd2014}).}
 \label{table:mediaeval results}
 \begin{tabular}{  l  c  c  c c} 
  
    \hline
    \textbf{Test Set} & \textbf{Precision} & \textbf{Recall} & \textbf{MAP@100} & \textbf{MAP2014} \\
    \hline
    YouTube-Gen & 49.7\% &  85.8\% & 86.0\% & 66.4\% \\
    Hollywood-Test & 41.1\% & 72.1\% & 72.7\% & 63.0\% \\
    \hline
    
 \end{tabular}
 \end{center}
 
\end{table}

\section{Discussion} \label{Discussion}

In this section, the results presented in \sref{Experiments and Results} are discussed. Before discussing the results of the Multi-Class and Binary classification tasks, the performance of the individual classifiers is 
discussed.  

\subsection{Individual Classifiers}

\begin{figure}[htbp]
\captionsetup{justification=justified,
singlelinecheck=false
}
\centering
\includegraphics[scale=1.0]{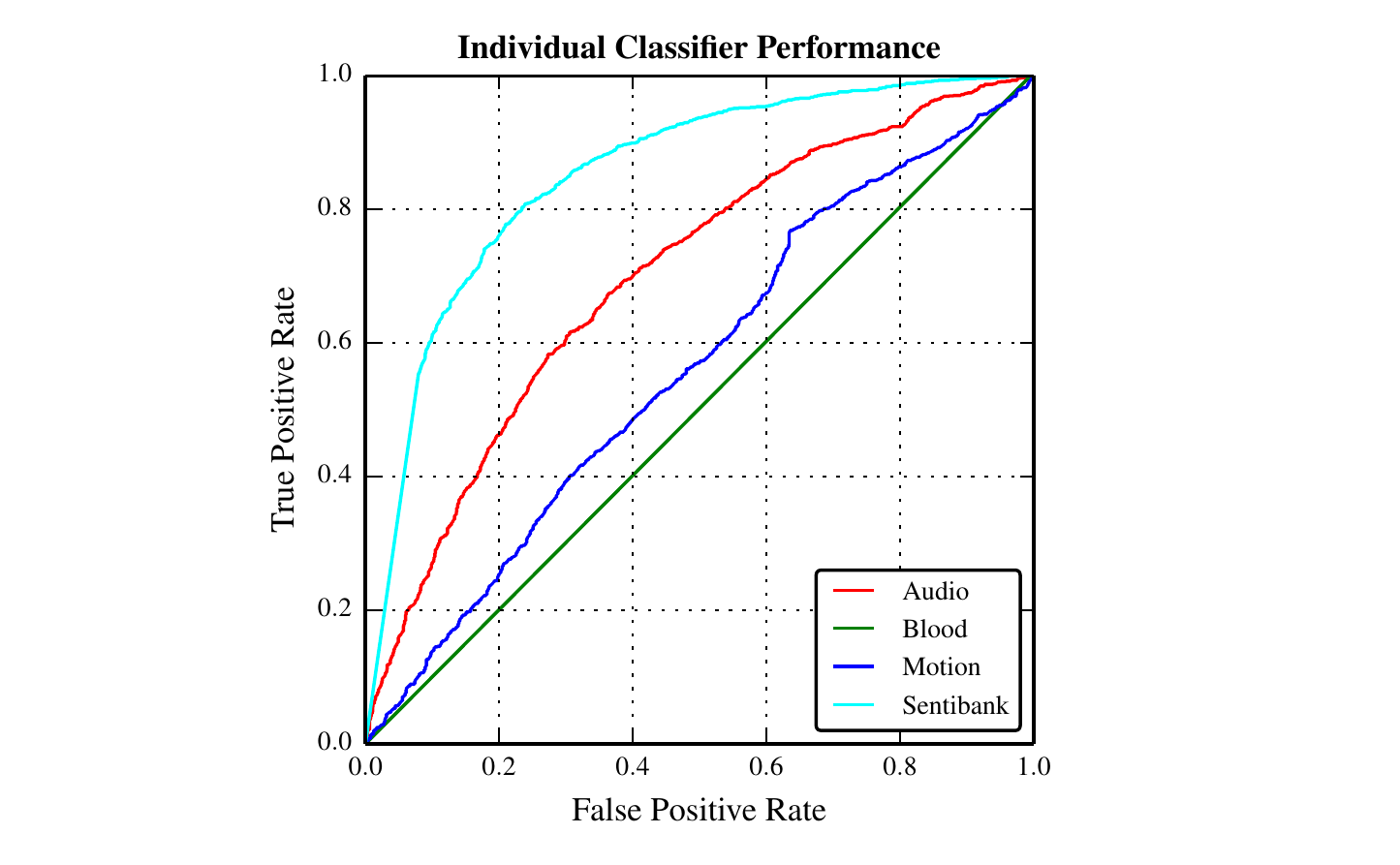}
\rule{\textwidth}{0.5pt}
\caption[Performance of individual binary classifiers on the test set.]{Performance of individual binary classifiers on the test set.}
\label{fig:classifiers}
\end{figure}

In both the classification tasks discussed in \sref{Experiments and Results}, a fusion of classifier scores is performed to get the final results. Hence, the performance of the system mainly depends on the individual 
performance of each of the classifiers and partially on the weights assigned to each of the classifiers. For the final classification results to be good, it is important that each of the classifiers have good individual 
performance. To get best performing classifiers, SVMs are trained using three different kernel functions (Linear, RBF, and Chi-Square) and the classifier with optimal performance on the test set are selected. Following
this approach, best performing classifiers for each feature type are selected. The performance of these selected classifiers on the test dataset in presented in \fref{fig:classifiers}. It can be observed that SentiBank 
and Audio are the two feature classifiers that show reasonable performance on the test set. Motion feature classifier has a performance which is a little better than chance and Blood has performance equivalent to chance.
A detailed discussion on the performance of each of these classifiers in the increasing order of their performance are presented next.

\subsubsection{Motion}

\begin{figure}
\captionsetup{justification=justified,
singlelinecheck=false
}
\begin{subfigure}{.5\textwidth}
  \includegraphics[scale=0.5]{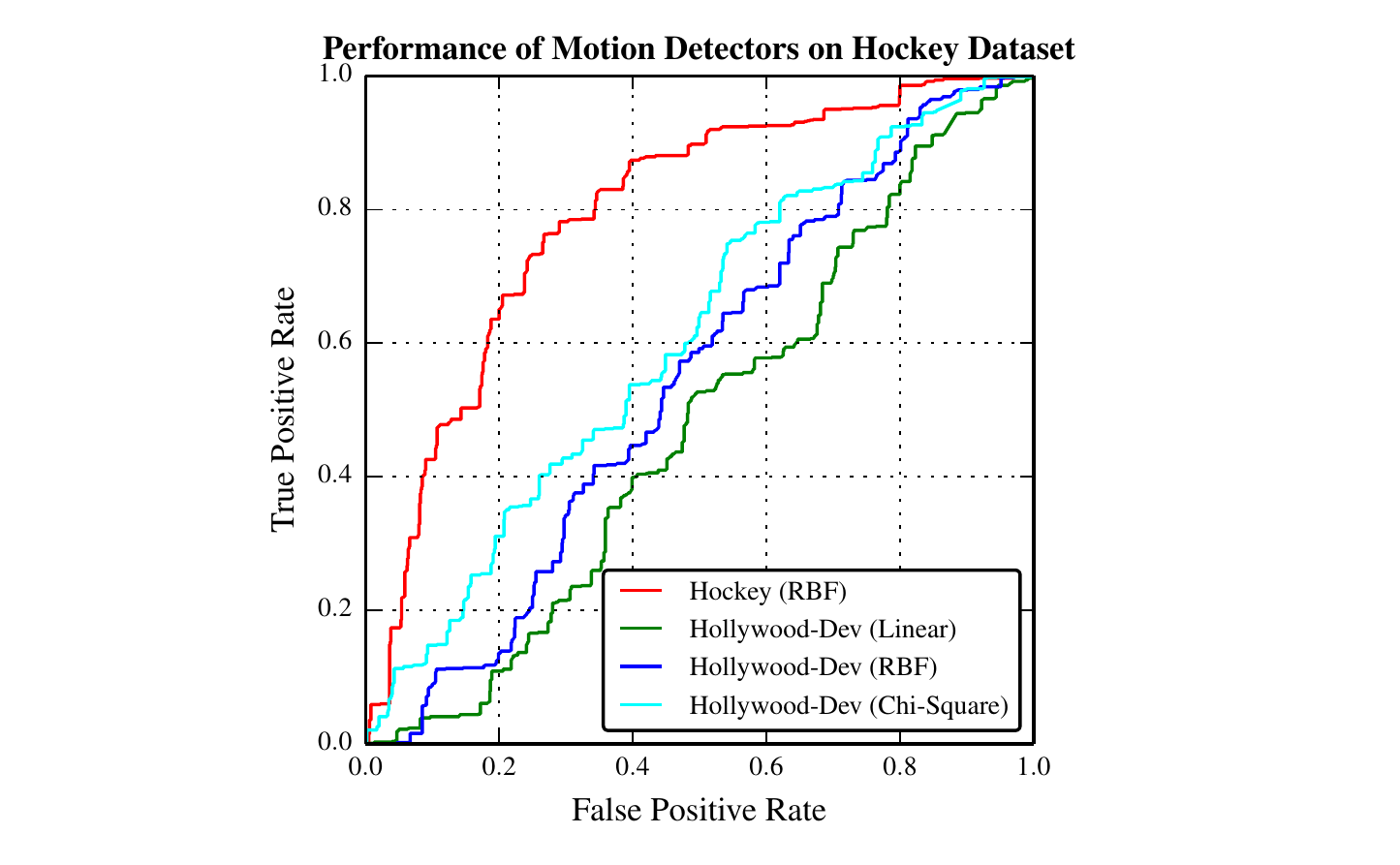}
  \label{fig:motion_hockey}
\end{subfigure}%
\begin{subfigure}{.5\textwidth}
  \includegraphics[scale=0.5]{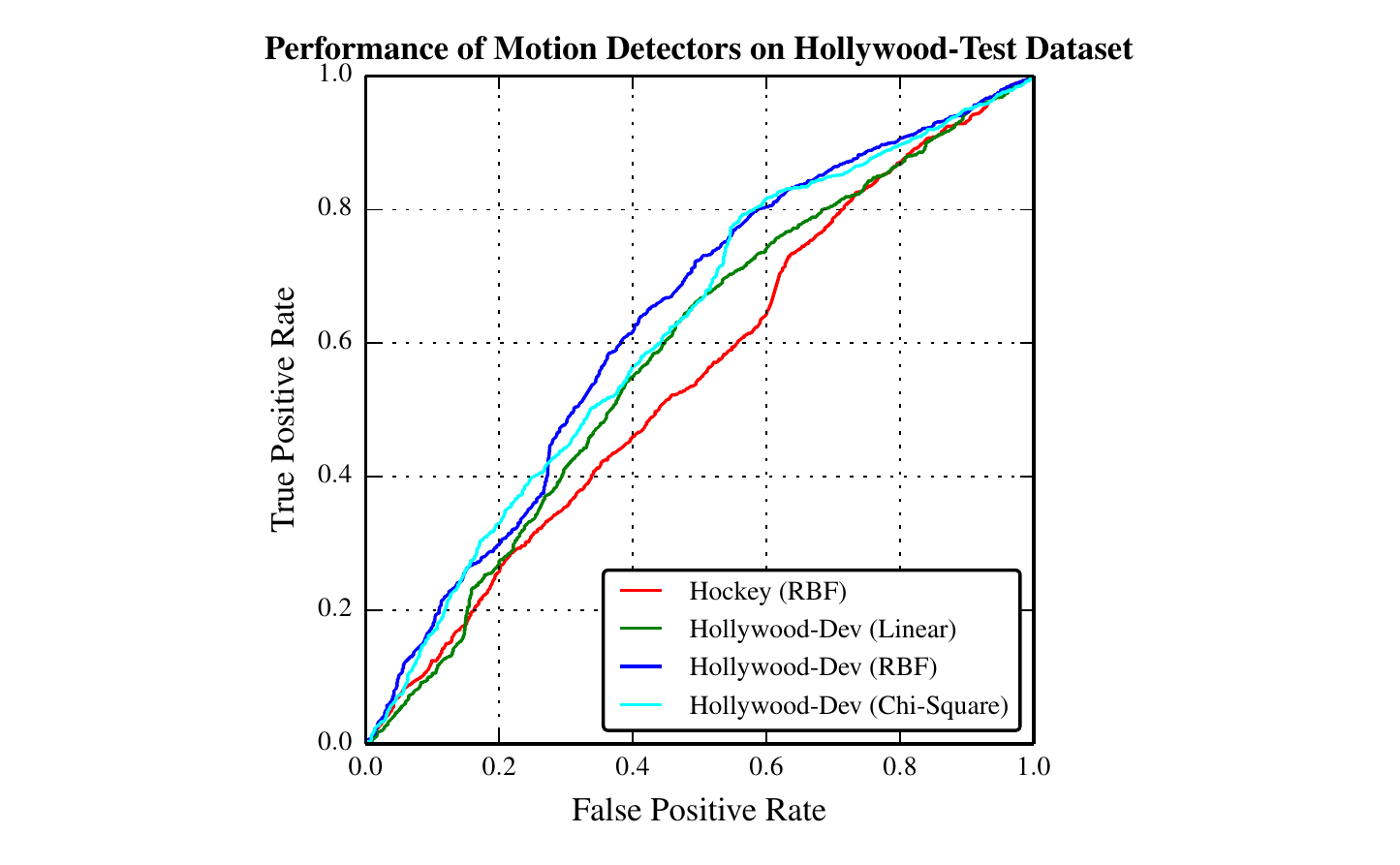}
  \label{fig:motion_hollywood}
\end{subfigure}
\rule{\textwidth}{0.5pt}
\caption[Performance of the Motion feature classifiers on Hockey and Hollywood-Test Datasets.]{Performance of the Motion feature classifiers on Hockey and Hollywood-Test Datasets. The red curve is for the classifier trained on 
the Hockey Dataset and the remaining three are for the three classifiers trained on Hollywood-Dev dataset with Linear, RBF and Chi-Square kernels.}
\label{fig:motion}
\end{figure}

As it is evident from  \fref{fig:classifiers}, the performance of the motion feature classifier on the test set is only a little better than chance. To understand the reason behind this, the performance of all the motion
feature classifiers, trained with different SVM kernels on available datasets are compared. Refer to \fref{fig:motion} for the comparison. In the figure, the left plot shows the performance of the classifiers on the test set
from Hockey dataset and the plot on the right shows the comparison on Hollywood-Test dataset. In both the graphs, the red curve corresponds to the classifier trained on the Hockey dataset and the remaining three curves 
correspond to classifiers trained on the Hollywood-Dev dataset. 

From both these plots, it can be observed that the performance of the classifiers trained and tested on the same dataset is reasonably good when compared to the classifiers which are trained on one dataset and tested on another.
In the plot on the left (TestSet: Hockey Dataset), the classifier trained on Hockey Dataset has better performance. Similarly, in the plot on the right (TestSet: Hollywood-Test), the performance of classifiers trained on 
Hollywood-Dev dataset have better performance. From there observations, it can be inferred that the motion feature representation learned from one dataset can not be transferred to another dataset. The reason for this could be
to the disparity in video resolution and video format between the datasets. The videos from the Hockey dataset and the Hollywood-Test dataset have different formats, and also, not all videos from Hollywood-Development and 
Hollywood-Test have the same format. The video format plays an important role as the procedure used to extract motion features (explained in \sref{Using Codec}) use motion information from video codecs. Length and resolution 
of a video will also have some effect, even though the procedure used here tries to reduce this by normalizing the extracted features with the length of the video segment and by aggregating the pixel motions over a pre-defined
number of sub-regions of the frame. Videos from Hockey dataset are very short segments of one second each and have small frame size and low quality. Whereas, the video segments from the Hollywood  dataset are longer and have 
larger frame size with better quality. One solution for this problem could be to convert all the videos to the same format, but even then there could be a problem due to improper video encoding. The other solution could be to 
use an Optical flow based approach to extract motion features (explained in \sref{Using Optical Flow}). But as explained earlier, this approach is tedious and may not work when there is blur due to motion in a video.

\subsubsection{Blood}

\begin{figure}[hbp]
  \centering
  \begin{subfigure}{0.30\linewidth}
    \centering
    \includegraphics[width=\linewidth]{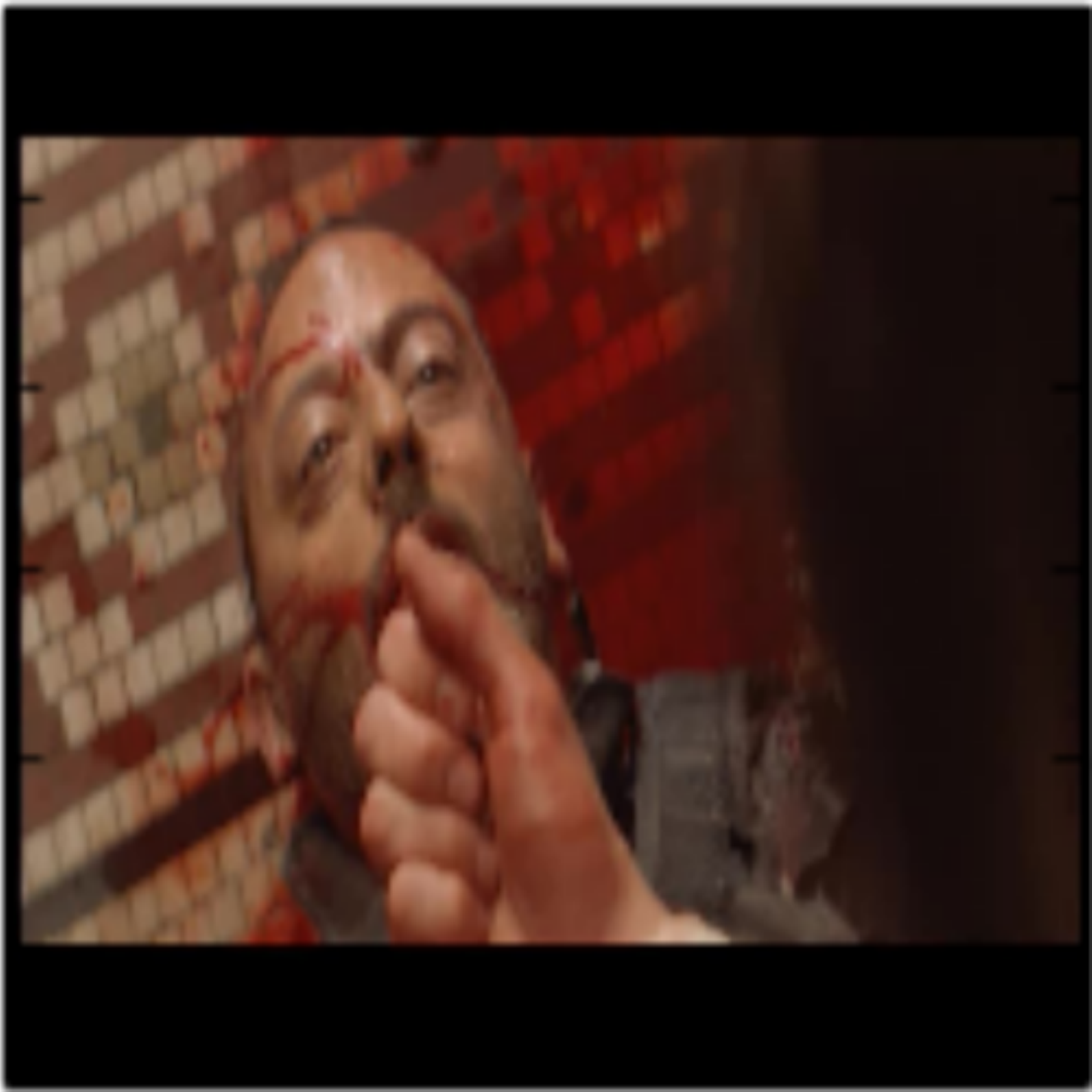}
    \caption{}
    \label{fig:img1}
  \end{subfigure}
  \begin{subfigure}{0.30\linewidth}
    \centering
    \includegraphics[width=\linewidth]{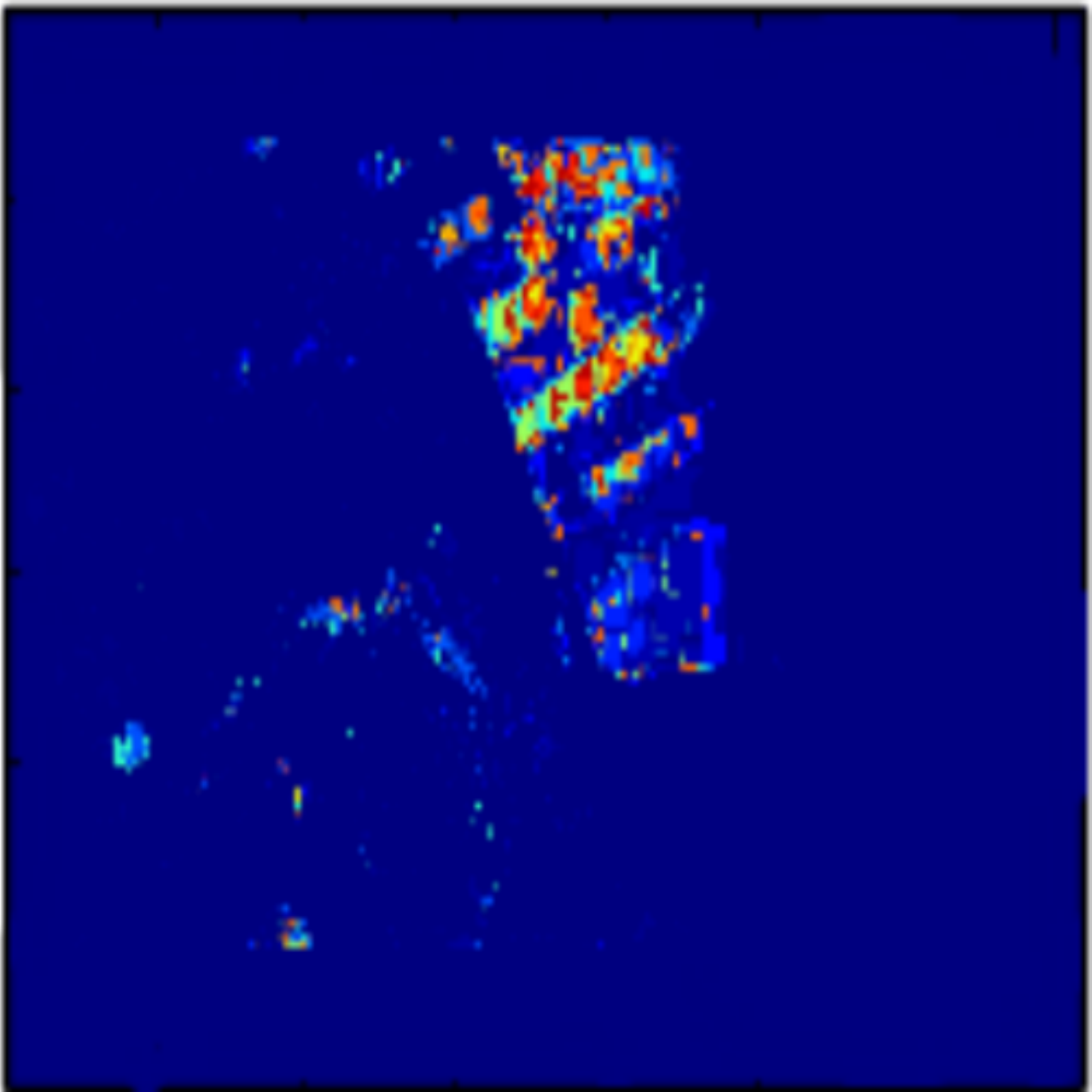}
    \caption{}
    \label{fig:img2}
  \end{subfigure}
  \begin{subfigure}{0.30\linewidth}
    \centering
    \includegraphics[width=\linewidth]{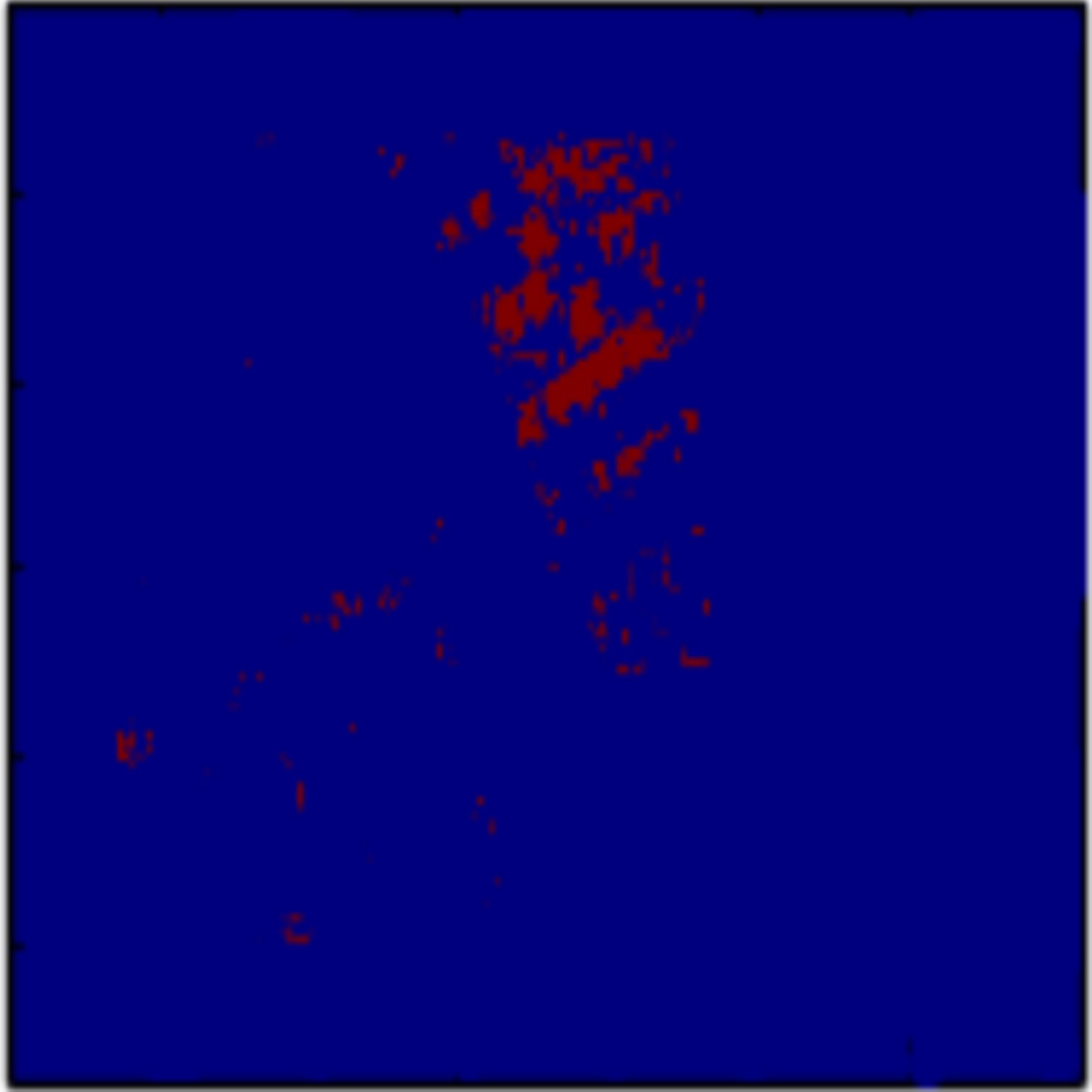}
    \caption{}
    \label{fig:img3}
  \end{subfigure}
  
  \par\bigskip
  
  \begin{subfigure}{0.30\linewidth}
    \centering
    \includegraphics[width=\linewidth]{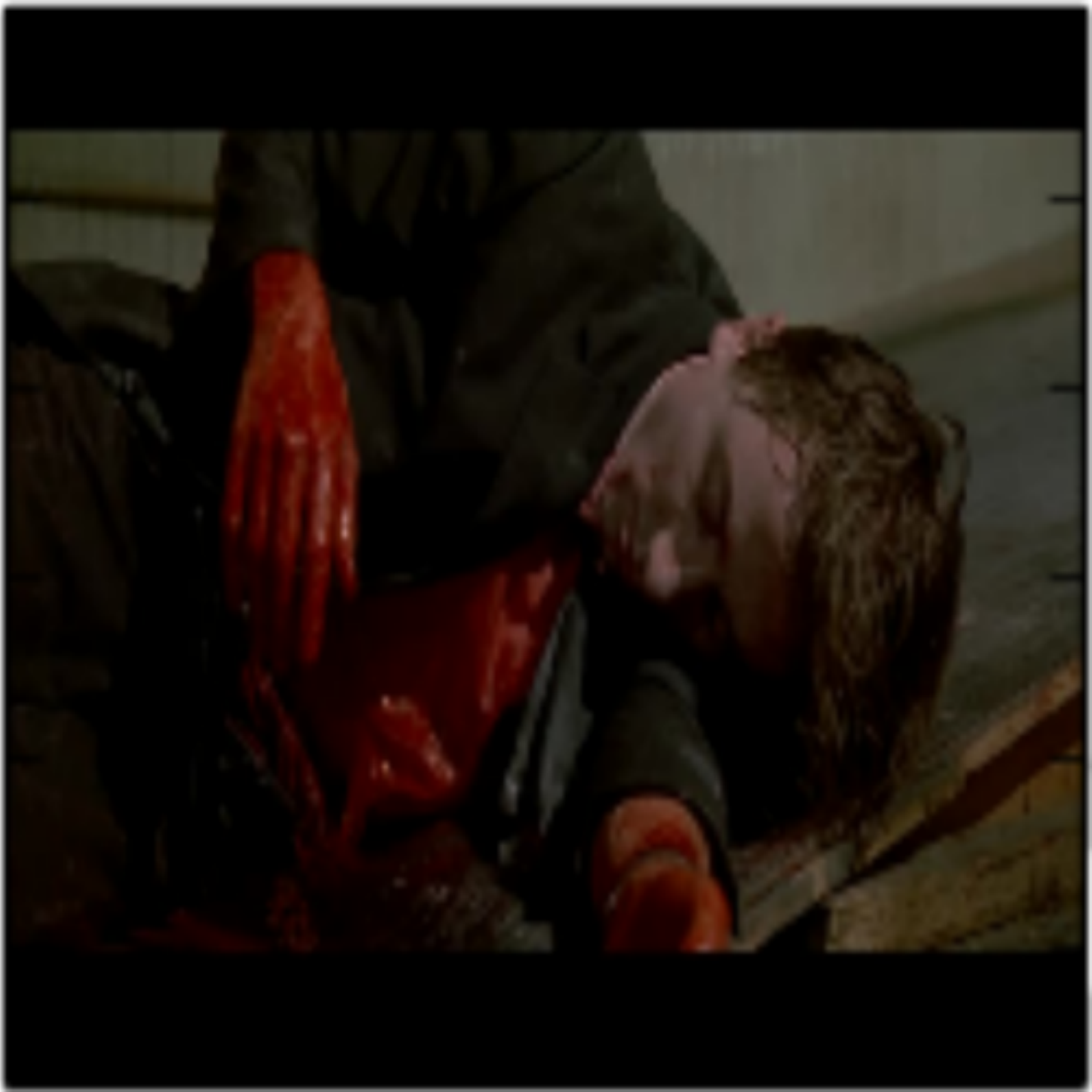}
    \caption{}
    \label{fig:img4}
  \end{subfigure}
  \begin{subfigure}{0.30\linewidth}
    \centering
    \includegraphics[width=\linewidth]{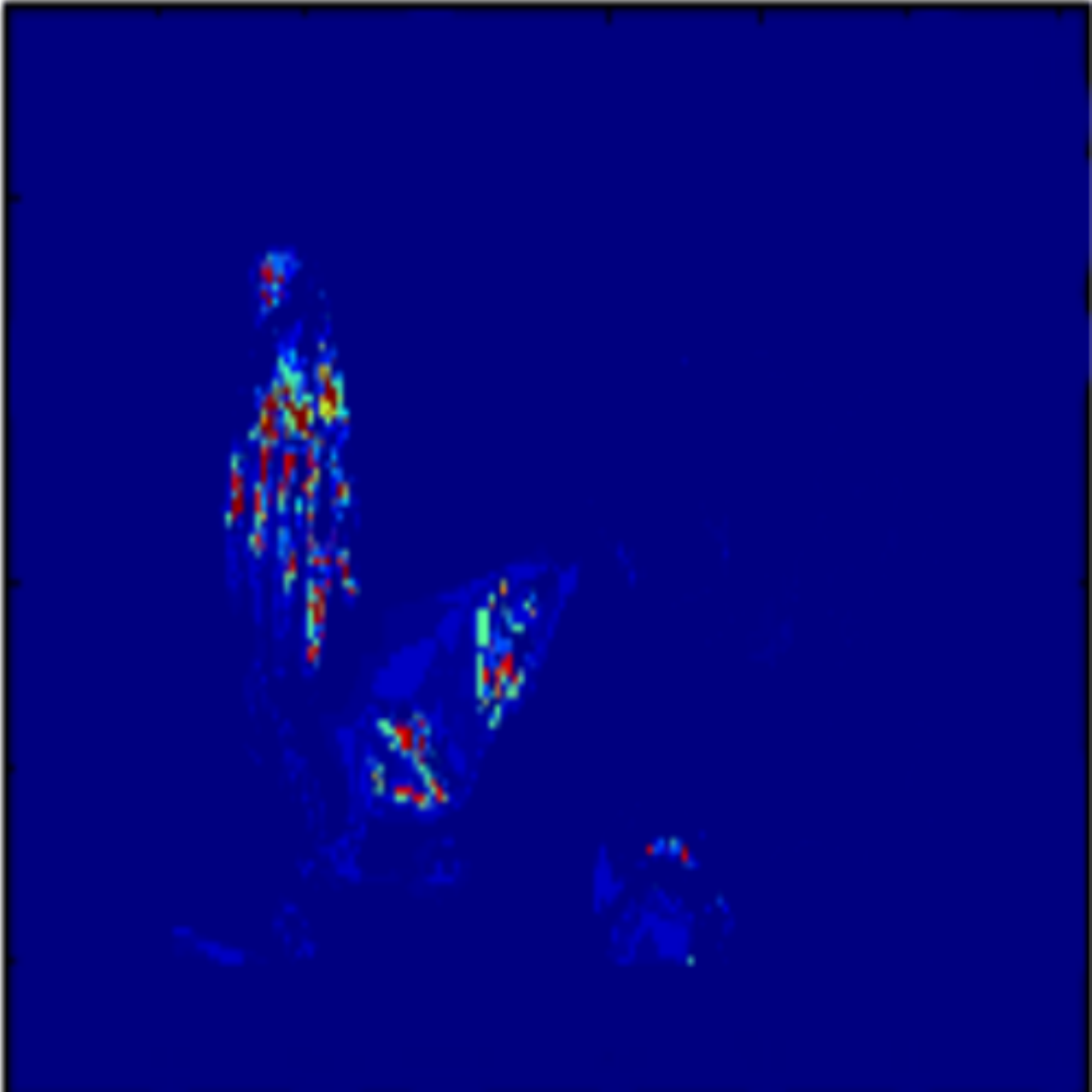}
    \caption{}
    \label{fig:img5}
  \end{subfigure}
  \begin{subfigure}{0.30\linewidth}
    \centering
    \includegraphics[width=\linewidth]{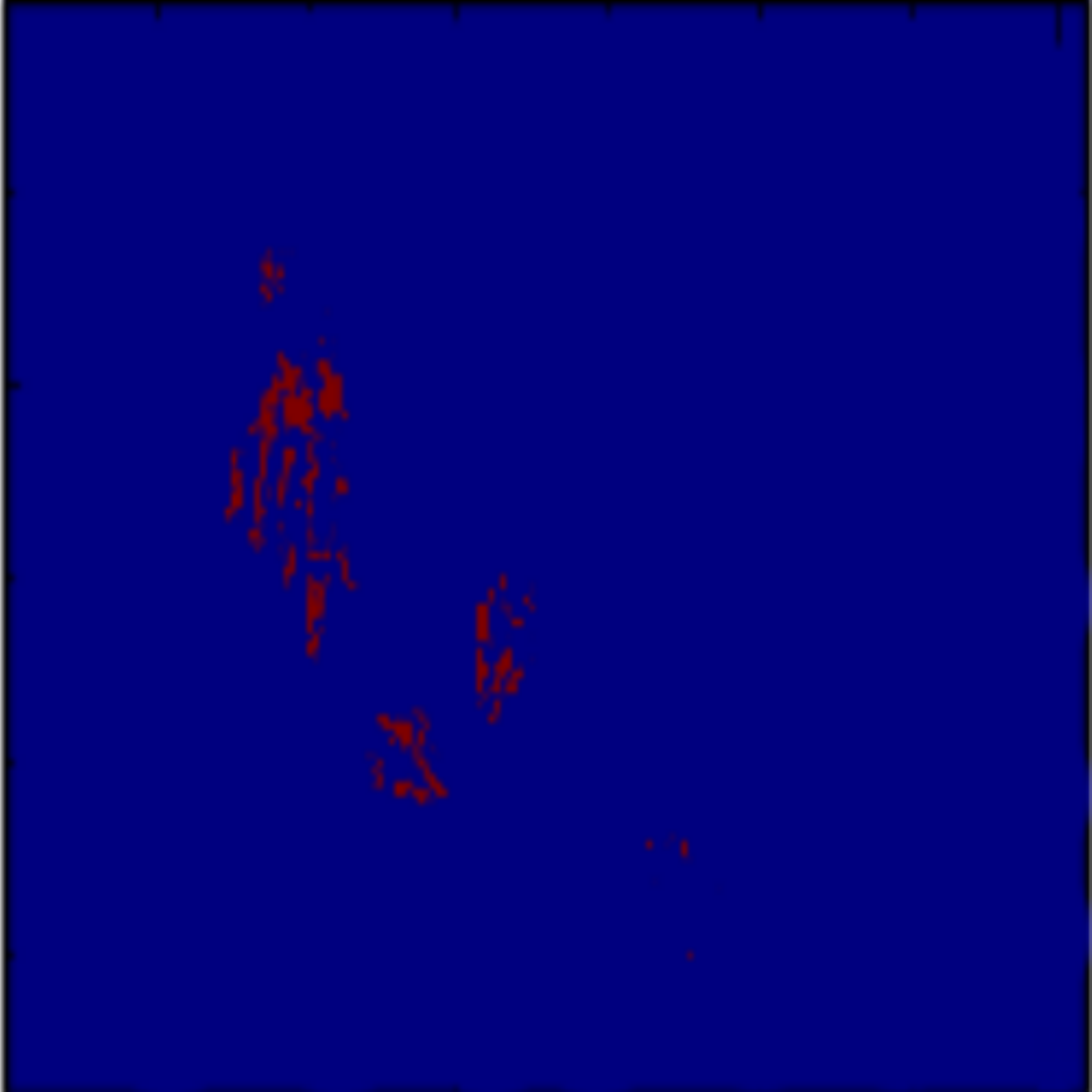}
    \caption{}
    \label{fig:img6}
  \end{subfigure}

  \par\bigskip  

  \rule{\textwidth}{0.5pt}
  \caption[Figure showing the performance of the blood detector on sample frames from the Hollywood dataset.]{Figure showing the performance of the blood detector on sample frames from the Hollywood dataset. The images in the 
  first column (A and D) are the input images, the second column images (B and E) are the blood probability maps and the images in the last column (C and F) are the binarized blood probability maps.}
  \label{fig:test_bloodmap}
\end{figure}

The performance of blood feature classifier on the test set is just as good as a chance. Refer to \fref{fig:classifiers} for the results. Here the problem is not with the feature extraction as the blood detector used for blood 
feature extraction has shown very good results in detecting regions containing blood in an image. Please refer to \fref{fig:bloodmap} for the performance of blood detector on images from the web and to \fref{fig:test_bloodmap}
for the performance of it on sample frames from the Hollywood dataset. From this, it is clear that the blood feature extractor is doing a pretty good job and it is not the problem with the feature extraction. Hence, it can be 
concluded that the problem is with the classifier training and it is due to the limited availability of training data.

In the VSD2014 dataset which is used for training, the video segments containing blood are annotated with labels (``Unnoticeable", ``Low", ``Medium", and ``High") representing the amount of blood contained in these segments.
There are very few segments in this dataset which are annotated with the label ``High", as a result of which, the SVM classifiers are unable to learn the feature representation of the frames containing blood effectively. The 
performance of this feature classifier can be improved by training it with a larger dataset with many instances of frames containing a high amount of blood. Alternatively images from Google can also be used to train this 
classifier.

\subsubsection{Audio}

Audio feature classifier is the second best-performing classifier (refer to \fref{fig:classifiers}) on the test set and this shows the importance of audio in violence detection. Although visual features are good indicators of
violent content, there are some scenes in which audio plays more important role. For example, scenes containing fights, gunshots, and explosions. These scenes have characteristic sounds and audio features such as MFCCs and 
Energy-entropy, can be used to detect sound patterns associated with these violent scenes.

In this work, MFCC features are used to describe audio content (refer to \sref{MFCC-Features}) as many previous works on violence detection (\citet{acar2011}, \citet{jiang2012}, \citet{lam2013}, etc.) have shown the 
effectiveness of MFCC features in detecting audio signatures associated with violent scenes. Other audio features such as energy entropy, pitch and power spectrum can also be used along with MFCC features to further improve
the performance of the feature classifier. But it is important to note that, audio alone is not sufficient to detect violence and it only plays an important role in detecting few violence classes such as Gunshots and Explosions
which have unique audio signatures.

\subsubsection{SentiBank}

\begin{figure}
\captionsetup{justification=justified,
singlelinecheck=false
}
\begin{subfigure}{1.0\textwidth}
  \includegraphics[scale=0.75]{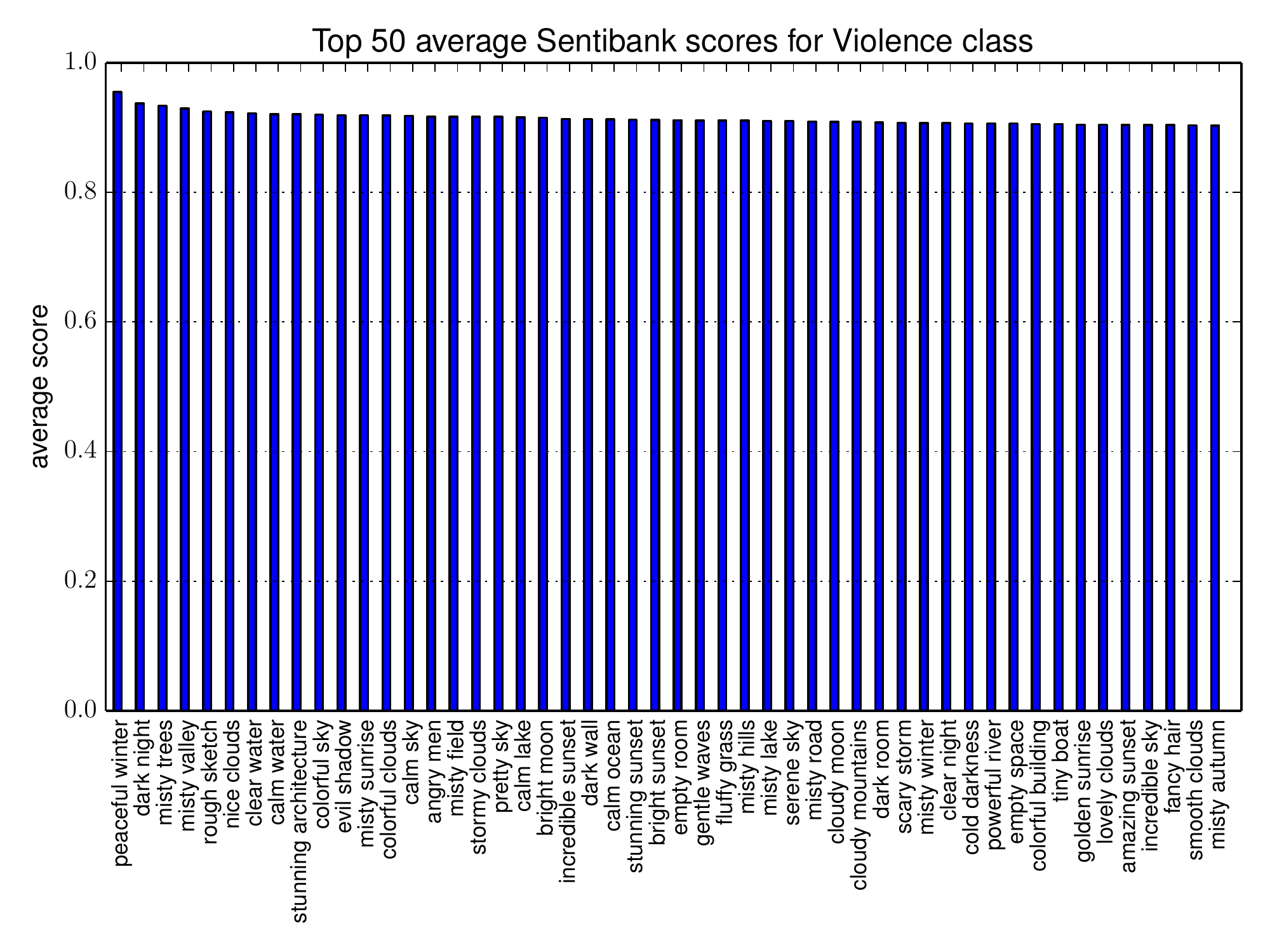}
  \label{fig:sentibank classes violence}
\end{subfigure}
\begin{subfigure}{1.0\textwidth}
 \includegraphics[scale=0.75]{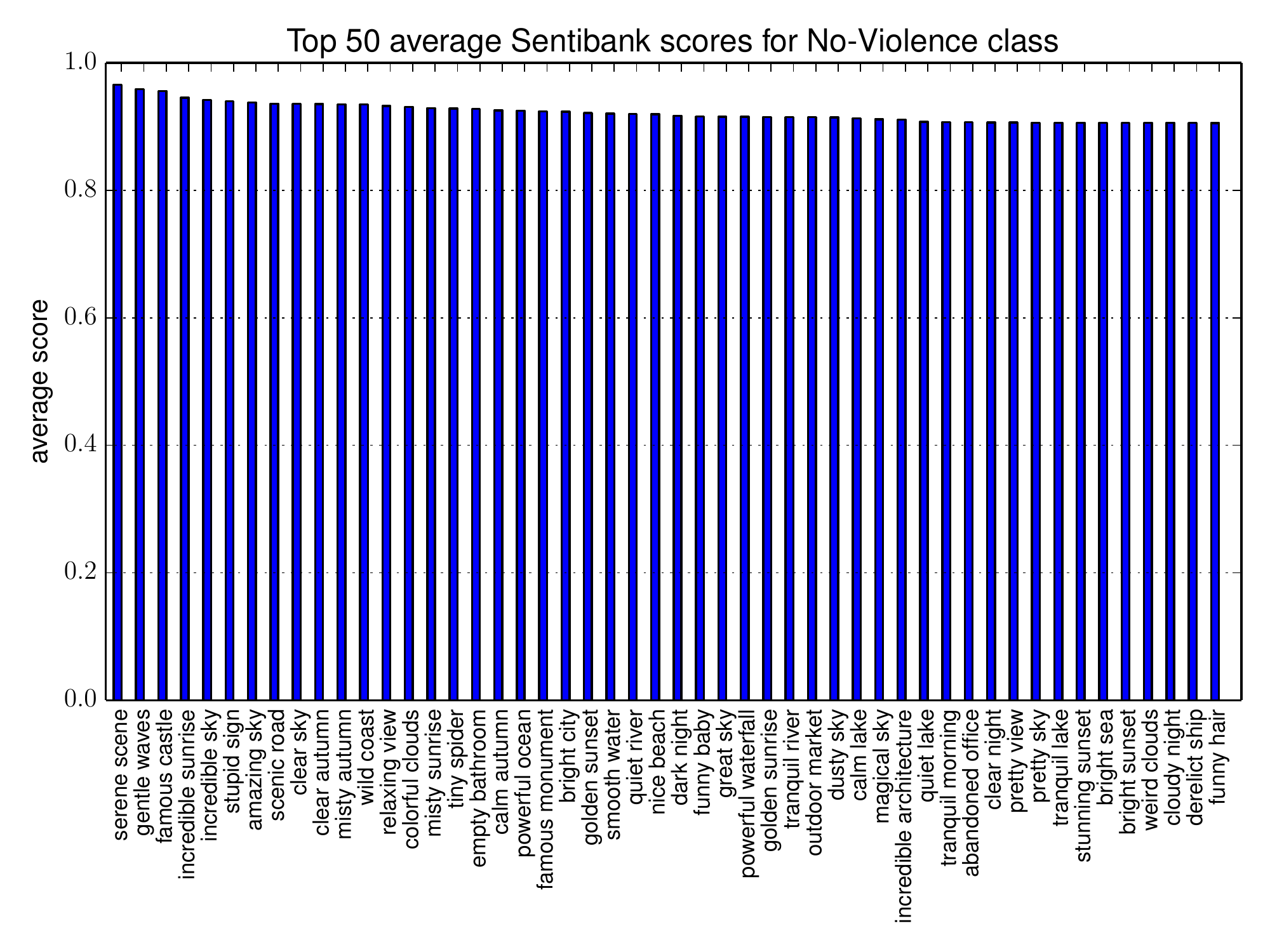}
  \label{fig:sentibank classes noviolence}
\end{subfigure}
\rule{\textwidth}{0.5pt}
\caption{Graphs showing average scores of Top 50 SentiBank ANPs for frames containing violence and no violence.}
\label{fig:sentibankclasses}
\end{figure}

\begin{figure}
\captionsetup{justification=justified,
singlelinecheck=false
}
\centering
\includegraphics[scale=0.5]{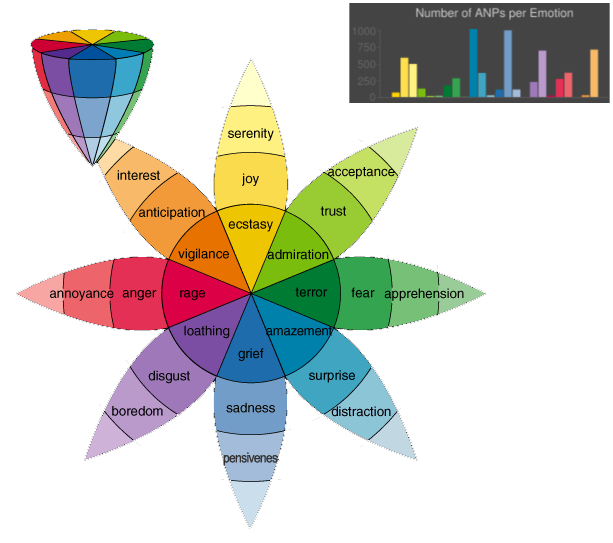}
\rule{\textwidth}{0.5pt}
\caption[Plutchik's wheel of emotions and the number of ANPs per emotion in VSO.]{Plutchik's wheel of emotions and the number of ANPs per emotion in VSO.}
\label{fig:emotions wheel ANPs}
\end{figure}

The SentiBank feature classifier has shown the best performance of all the feature classifiers (Refer to \fref{fig:classifiers}) and has contributed strongly to the overall performance of the system. This demonstrates the 
power of SentiBank, in detecting complex visual sentiments such as violence. \fref{fig:sentibankclasses} shows the average scores for top \num{50} ANPs for frames containing violence and no violence. As it can be observed the
list of ANPs with highest average scores for violence and no-violence class are very different and this is the reason behind the very good performance of SentiBank in separating violence class from no-violence class. Note that,
not all the adjectives in the ANP list for violence class describe violence. This could be due to many different reasons, one of which could be the fact that, of the \num{1200} ANPs used in SentiBank only a few describe the
emotions related to violence (like fear, terror, rage, anger etc.,). Please refer to \fref{fig:emotions wheel ANPs} which shows the Plutchik’s Wheel of Emotions and the distribution of ANPs for each category of emotion in VSO.

\subsection{Fusion Weights}

As mentioned earlier (\sref{Feature Fusion}), the final classification scores are calculated by late fusion of individual classifier scores using weighted sum approach. The weights used here are calculated using a grid-search 
approach with the goal to minimize the Equal Error Rate (EER). So, weights play an important role in determining the overall classification performance of the system. Note that all these weights are calculated on the test set.
In \tref{table:weights}, the weights of the classifiers for each of the eight violence class, obtained using the grid-search technique, are presented. From the weights obtained, the following observations about weight 
distribution can be made, (i) For most of the violence classes, the highest weight is assigned to SentiBank as it is the most discriminative feature. (ii) Audio has received the highest weight for violence classes such as 
Gunshots, Explosions, and Fights where audio plays a very important role. (iii) Blood has received high weights for violence classes such as Screams, Gunshots, and Firearms. This is interesting as a video segment belonging to 
any of these violence classes can also have blood in it. (iv) Motion has received the least weight in most of the violence classes as it the least performing feature. But, it can also be observed that it has a higher weight 
for the class Fights where a lot of motion can be expected.

If the weights assigned to each of the violence classes are analyzed the following observations can be made, (i) For the class Gunshots, the highest distribution weights is between Audio (\num{0.5}) and Blood (\num{0.45}).
This is expected as audio features play an important role in detecting gunshots and the scenes containing gunshots are also expected to have a lot of blood. (ii) Audio (\num{0.4}), and visual features (Motion - \num{0.25} and
SentiBank - \num{0.30}) have received an almost equal amount of weight for the class Fights. This is expected as both audio and visual features are important in detecting scenes containing fights. (iii) For the class Explosions,
highest weights are assigned to Audio (\num{0.9}) which is expected, as audio features are crucial in detecting explosions. (iv) Fire is a violence class where visual features are expected to have high weights and as expected
the best performing visual feature, SentiBank (\num{0.85}), is assigned the highest weight. (v) Violence class Cold arms contain scenes which have the presence of any cold weapon (e.g., knives, swords, arrows, halberds, etc.).
For this class, visual features are expected to have high weights. And as expected, SentiBank (\num{0.95}) has the highest weight for this class. (vi) ``Firearms" is the violence class in which the scenes contain guns and 
firearms. Similar to the above class, visual features are expected to have high weights. For this class, SentiBank (\num{0.6}) and Blood (\num{0.3}) have received the highest distribution of weights. The reason for Blood being
assigned a higher weight might be due to the fact that most of the scenes containing guns will also contain bloodshed. (vii) For the class Blood, the feature Blood is expected to have the highest weight. But feature Blood 
(\num{0.05}) received only a small weight and SentiBank (\num{0.95}) gained the highest weight. This is not an expected result and this could be due to the poor performance of the Blood feature classifier on the test set.
(viii) It is intuitive to expect Audio to have higher weights for class ``Screams" as audio features play an important role in detecting screams. But, the weights obtained here are against this intuition. Audio has received very
less weight whereas SentiBank has received highest weight. Overall, the weights obtained from the grid-search are more or less as expected for most of the classes. Better weight distribution could be obtained if the
performance of individual classifiers on the test is improved.

\subsection{Multi-Class Classification}

In this section, the results obtained in the multi-class classification task are discussed. Please refer to \fref{fig:multiclass} for the results obtained in this task. From the figure, the following observations can be drawn
(i) The system shows good performance (EER of around 30\%) in detecting Gunshots. (ii) For the violence classes, Cold arms, Blood and Explosions, the system shows moderate performance (EER of around 40\%). (iii) For the 
remaining violence classes (Fights, Screams, Fire, Firearms) the performance is as good as a chance (EER of more than 45\%). These results suggest that there is huge scope for improvement, but, it is important to remember that
violence detection is not a trivial task and distinguishing between different classes of violence is, even more, difficult. All the approaches proposed so far have only concentrated on detecting the presence or absence of 
violence, but not on detecting the category of violence. The novel approach proposed in this work is one of the first in this direction and there are no baseline systems to compare the performance with. The results obtained
from this work will serve as a baseline for future works in this area.

In this system, the late fusion approach is followed which has shown good results in a similar multimedia concept detection task of adult content detection (\citet{schulze2014}). Hence, the poor performance of the system can 
not be attributed to the approach followed. The performance of the system depends on the performance of individual classifiers and the fusion weight assigned to them for each of the violence classes. As the fusion weights are
adjusted to minimize the EER using the Grid-Search technique, the overall performance of the system solely depends on the performance of the individual classifiers. So, to improve the performance of the system in this task,
it is necessary to improve the performance of individual classifiers in detecting violence.

\subsection{Binary Classification}

The results for the binary classification task are presented in \fref{fig:binary}. This task is an extension to the multi-class classification task. As explained earlier, in this task, a video segment is categorized as ``Violence" 
if the output probability for any one of the violence classes is more than the threshold of \num{0.5}. The performance of the system in this task is evaluated on two datasets, Hollywood-Test, and YouTube-Generalization. It can
be observed that the performance of the system on these datasets is a little better than chance. It can also be observed that the performance is better on Hollywood-Test dataset than YouTube-Generalization 
dataset. This is expected as all the classifiers are trained on data from Hollywood-Development dataset which have similar video content to that of Hollywood-Test dataset. The precision, recall and accuracy values obtained by
the system for this task are presented in \tref{table:results}. The results obtained by the best performing team in this task from MediaEval-2014 are given in \tref{table:mediaeval results}. 

These results can not be directly compared, even though the same dataset is used, as the process used for evaluation is not the same. In MediaEval-2014, a system is expected to output the start and end frame for the video
segments which contain violence and, if the overlap between the ground truth and the output frame intervals is more than \num{50}\% then it is considered as a hit. Please refer to \citet{schedl6vsd2014} for more information on
the process followed in MediaEval-2014. In the proposed approach, the system categorizes each \num{1}-second segment of the input video to be of class ``Violence" or ``No Violence" and the system performance is calculated by
comparing this with the ground truth. This evaluation criteria used here is much more stringent and more granular when compared to the one used in MediaEval-2014. Here, as the classification is done for each \num{1}-second
segment, there is no need for a strategy to penalize detection of shorter segments. MAP metric is used  for selecting the best performing system in MediaEval whereas, in the proposed system, the EER of the system is optimized. 

Even though the results obtained from this system can not be directly compared to the MediaEval results, it can be observed that the performance of this system is comparable to, if not better than, the best performing system
from MediaEval-2014 even though strict evaluation criteria are used. These results suggest that the system developed using the proposed novel approach is better than the existing state-of-art systems in this area of violence 
detection.

\section{Summary}

In this chapter, a detailed discussion on the evaluation of the developed system is presented. In the \sref{Datasets}, details of the datasets used in this work are explained and in the next section \sref{Setup}, the experimental
setup is discussed. In \sref{Experiments and Results} the experiments and their results are presented, followed by a detailed discussion on the obtained results in \sref{Discussion}.

\afterpage{\blankpage}
\clearpage

\chapter{Conclusions and Future Work} 

\label{Chapter5}

\lhead{Chapter 5. \emph{Conclusions and Future Work} } 

In this chapter, the conclusions and the directions in which the existing work can be extended are discussed in the \sref{Conclusions} and \sref{Future Work} respectively. 

\section{Conclusions} \label{Conclusions}

In this work, an attempt has been made to develop a system to detect violent content in videos using both visual and audio features. Even though the approach used in this work is motivated by the earlier works in this area, 
the following are the unique aspects of it: (i) Detection of different classes of violence, (ii) the use of SentiBank feature to describe visual content of a video, (iii) the Blood detector and the blood
model developed using images from the web, and (iv) using information from video codec to generate motion features. Here is a brief overview of the process used to develop this system.

As violence is not a physical entity, the detection of it in a video is not a trivial task. Violence is a visual concept and to detect it there is a need to use multiple features. In this work, MFCC features were used to 
describe audio content and Blood, Motion and SentiBank features are used to describe visual content. SVM classifiers were trained for each of the selected features and the individual classifier scores were combined by weighted
sum to get the final classification scores for each of the violence classes. The weights for each class are found using a grid-search approach with the optimizing criteria to be the minimum EER. Different datasets are used
in this work, but the most important one is the VSD dataset, which is used for training the classifiers, calculating the classifier weights and for testing the system.

The performance of the system is evaluated on two different classification tasks, Multi-Class, and Binary classification. In Multi-Class classification task, the system has to detect the class of violence present in a video
segment. This is a much more difficult task than just detecting the presence of violence and the system presented here is one of the first to tackle this problem. The Binary classification task is where the system has to just
detect the presence of violence without having to find the class of violence. In this task, if the final classification score from the Multi-Class classification task for any of the violence class is more than \num{0.5}, then
the video segment is categorized as ``Violence" else, it is categorized as ``No Violence". The results from the Multi-Class classification task is far from perfect and there is room for improvement, whereas, the results on the 
Binary classification tasks are better than the existing benchmark results from MediaEval-2014. However, these results are definitely encouraging. In \sref{Future Work}, a detailed discussion on the possible
directions in which the current work can be extended are presented.

\section{Future Work} \label{Future Work}

There are many possible directions in which the current work can be extended. One direction would be to improve the performance of the existing system. For that, the performance of the individual classifiers has to be improved.
Motion and Blood are the two features whose classifier performance needs resonable improvement. As explained in \sref{Discussion}, the approach used to extract motion features has to be changed for improving the performance of 
the motion classifier. For Blood, the problem is with the dataset used for training the classifier but not the feature extractor. An appropriate dataset with decent amount of frames containing blood should be used for training.
Making these improvements should be the first step towards building a better system. Another direction for the future work would be to adapt this system and develop different tools for different applications. For example, 
(i) a tool could be developed which could extract the video segments containing violence from a given input video. This could be helpful in video tagging. (ii) A similar tool could be developed for parental control where the 
system could be used to rate a movie depending on the amount of violent content in it. Another possible direction for future work is, to improve the speed of the system so that it can be used in the real-time detection of 
violence from the video feed of security cameras. The improvements needed for developing such a system will not be trivial.


\label{Bibliography}

\lhead{\emph{Bibliography}} 

\bibliographystyle{abbrvnat} 
\bibliography{Bibliography} 

\begin{thebibliography}{57}
\providecommand{\natexlab}[1]{#1}
\providecommand{\url}[1]{\texttt{#1}}
\expandafter\ifx\csname urlstyle\endcsname\relax
  \providecommand{\doi}[1]{doi: #1}\else
  \providecommand{\doi}{doi: \begingroup \urlstyle{rm}\Url}\fi

\bibitem[Acar et~al.(2011)Acar, Spiegel, Albayrak, and Labor]{acar2011}
E.~Acar, S.~Spiegel, S.~Albayrak, and D.~Labor.
\newblock Mediaeval 2011 affect task: Violent scene detection combining audio
  and visual features with svm.
\newblock In \emph{MediaEval}, 2011.

\bibitem[Blake and Shiffrar(2007)]{blake2007}
R.~Blake and M.~Shiffrar.
\newblock Perception of human motion.
\newblock \emph{Annu. Rev. Psychol.}, 58:\penalty0 47--73, 2007.

\bibitem[Blog-FB(2015)]{facebookstats}
Blog-FB.
\newblock Facebook statistics, 2015.
\newblock URL
  \url{http://media.fb.com/2015/01/07/what-the-shift-to-video-means-for-creators/}.
\newblock Online: accessed 08-Jan-2016.

\bibitem[Borth et~al.(2013)Borth, Ji, Chen, Breuel, and Chang]{borth2013}
D.~Borth, R.~Ji, T.~Chen, T.~Breuel, and S.-F. Chang.
\newblock Large-scale visual sentiment ontology and detectors using adjective
  noun pairs.
\newblock In \emph{Proceedings of the 21st ACM international conference on
  Multimedia}, pages 223--232. ACM, 2013.

\bibitem[Bradski(2000)]{opencv}
G.~Bradski.
\newblock Opencv.
\newblock \emph{Dr. Dobb's Journal of Software Tools}, 2000.

\bibitem[Bushman and Huesmann(2006)]{bushman2006}
B.~J. Bushman and L.~R. Huesmann.
\newblock Short-term and long-term effects of violent media on aggression in
  children and adults.
\newblock \emph{Archives of Pediatrics \& Adolescent Medicine}, 160\penalty0
  (4):\penalty0 348--352, 2006.

\bibitem[Cai et~al.(2006)Cai, Lu, Hanjalic, Zhang, and Cai]{cai2006}
L.-H. Cai, L.~Lu, A.~Hanjalic, H.-J. Zhang, and L.-H. Cai.
\newblock A flexible framework for key audio effects detection and auditory
  context inference.
\newblock \emph{Audio, Speech, and Language Processing, IEEE Transactions on},
  14\penalty0 (3):\penalty0 1026--1039, 2006.

\bibitem[Chan et~al.(1999)Chan, Harvey, and Smith]{chan1999}
Y.~Chan, R.~Harvey, and D.~Smith.
\newblock Building systems to block pornography.
\newblock In \emph{Challenge of Image Retrieval, BCS Electronic Workshops in
  Computing series}, pages 34--40, 1999.

\bibitem[Chang and Lin(2011)]{Chang2011}
C.-C. Chang and C.-J. Lin.
\newblock {LIBSVM}: A library for support vector machines.
\newblock \emph{ACM Transactions on Intelligent Systems and Technology},
  2:\penalty0 27:1--27:27, 2011.
\newblock Software available at \url{http://www.csie.ntu.edu.tw/~cjlin/libsvm}.

\bibitem[Chen and Hauptmann(2009)]{chen2009}
M.-y. Chen and A.~Hauptmann.
\newblock Mosift: Recognizing human actions in surveillance videos.
\newblock 2009.

\bibitem[Cheng et~al.(2003)Cheng, Chu, and Wu]{cheng2003}
W.-H. Cheng, W.-T. Chu, and J.-L. Wu.
\newblock Semantic context detection based on hierarchical audio models.
\newblock In \emph{Proceedings of the 5th ACM SIGMM international workshop on
  Multimedia information retrieval}, pages 109--115. ACM, 2003.

\bibitem[Clarin et~al.(2005)Clarin, Dionisio, Echavez, and Naval]{clarin2005}
C.~Clarin, J.~Dionisio, M.~Echavez, and P.~Naval.
\newblock Dove: Detection of movie violence using motion intensity analysis on
  skin and blood.
\newblock \emph{PCSC}, 6:\penalty0 150--156, 2005.

\bibitem[Clarke et~al.(2005)Clarke, Bradshaw, Field, Hampson, Rose,
  et~al.]{clarke2005}
T.~J. Clarke, M.~F. Bradshaw, D.~T. Field, S.~E. Hampson, D.~Rose, et~al.
\newblock The perception of emotion from body movement in point-light displays
  of interpersonal dialogue.
\newblock \emph{Perception-London}, 34\penalty0 (10):\penalty0 1171--1180,
  2005.

\bibitem[Datta et~al.(2002)Datta, Shah, and Lobo]{datta2002}
A.~Datta, M.~Shah, and N.~D.~V. Lobo.
\newblock Person-on-person violence detection in video data.
\newblock In \emph{Pattern Recognition, 2002. Proceedings. 16th International
  Conference on}, volume~1, pages 433--438. IEEE, 2002.

\bibitem[Demarty et~al.(2010)Demarty, Penet, Gravier, and
  Soleymani]{demarty2011}
C.-H. Demarty, C.~Penet, G.~Gravier, and M.~Soleymani.
\newblock The mediaeval 2011 affect task.
\newblock 2010.

\bibitem[Demarty et~al.(2012)Demarty, Penet, Gravier, and
  Soleymani]{demarty2012}
C.-H. Demarty, C.~Penet, G.~Gravier, and M.~Soleymani.
\newblock A benchmarking campaign for the multimodal detection of violent
  scenes in movies.
\newblock In \emph{Computer Vision--ECCV 2012. Workshops and Demonstrations},
  pages 416--425. Springer, 2012.

\bibitem[Demarty et~al.(2014{\natexlab{a}})Demarty, Ionescu, Jiang, Quang,
  Schedl, and Penet]{demarty2014benchmarking}
C.-H. Demarty, B.~Ionescu, Y.-G. Jiang, V.~L. Quang, M.~Schedl, and C.~Penet.
\newblock Benchmarking violent scenes detection in movies.
\newblock In \emph{Content-Based Multimedia Indexing (CBMI), 2014 12th
  International Workshop on}, pages 1--6. IEEE, 2014{\natexlab{a}}.

\bibitem[Demarty et~al.(2014{\natexlab{b}})Demarty, Penet, Ionescu, Gravier,
  and Soleymani]{demarty2014multimodal}
C.-H. Demarty, C.~Penet, B.~Ionescu, G.~Gravier, and M.~Soleymani.
\newblock Multimodal violence detection in hollywood movies: State-of-the-art
  and benchmarking.
\newblock In \emph{Fusion in Computer Vision}, pages 185--208. Springer,
  2014{\natexlab{b}}.

\bibitem[Demarty et~al.(2014{\natexlab{c}})Demarty, Penet, Soleymani, and
  Gravier]{demarty2014}
C.-H. Demarty, C.~Penet, M.~Soleymani, and G.~Gravier.
\newblock Vsd, a public dataset for the detection of violent scenes in movies:
  design, annotation, analysis and evaluation.
\newblock \emph{Multimedia Tools and Applications}, pages 1--26,
  2014{\natexlab{c}}.

\bibitem[Dempster et~al.(1977)Dempster, Laird, and Rubin]{dempster1977}
A.~P. Dempster, N.~M. Laird, and D.~B. Rubin.
\newblock Maximum likelihood from incomplete data via the em algorithm.
\newblock \emph{Journal of the royal statistical society. Series B
  (methodological)}, pages 1--38, 1977.

\bibitem[Deniz et~al.(2014)Deniz, Serrano, Bueno, and Kim]{deniz2014}
O.~Deniz, I.~Serrano, G.~Bueno, and T.~Kim.
\newblock Fast violence detection in video.
\newblock In \emph{The 9th International Conference on Computer Vision Theory
  and Applications (VISAPP)}, 2014.

\bibitem[Eyben and Schuller(2015)]{eyben2015}
F.~Eyben and B.~Schuller.
\newblock opensmile:): the munich open-source large-scale multimedia feature
  extractor.
\newblock \emph{ACM SIGMultimedia Records}, 6\penalty0 (4):\penalty0 4--13,
  2015.

\bibitem[Eyben et~al.(2013)Eyben, Weninger, Lehment, Schuller, and
  Rigoll]{eyben2013}
F.~Eyben, F.~Weninger, N.~Lehment, B.~Schuller, and G.~Rigoll.
\newblock Affective video retrieval: Violence detection in hollywood movies by
  large-scale segmental feature extraction.
\newblock \emph{PloS one}, 8\penalty0 (12):\penalty0 e78506, 2013.

\bibitem[Farneb{\"a}ck(2003)]{farneback2003}
G.~Farneb{\"a}ck.
\newblock Two-frame motion estimation based on polynomial expansion.
\newblock In \emph{Image Analysis}, pages 363--370. Springer, 2003.

\bibitem[Flood(2009)]{flood2009}
M.~Flood.
\newblock The harms of pornography exposure among children and young people.
\newblock \emph{Child abuse review}, 18\penalty0 (6):\penalty0 384--400, 2009.

\bibitem[Gill et~al.(2007)Gill, Arlitt, Li, and Mahanti]{gill2007}
P.~Gill, M.~Arlitt, Z.~Li, and A.~Mahanti.
\newblock Youtube traffic characterization: a view from the edge.
\newblock In \emph{Proceedings of the 7th ACM SIGCOMM conference on Internet
  measurement}, pages 15--28. ACM, 2007.

\bibitem[Gong et~al.(2008)Gong, Wang, Jiang, Huang, and Gao]{gong2008}
Y.~Gong, W.~Wang, S.~Jiang, Q.~Huang, and W.~Gao.
\newblock Detecting violent scenes in movies by auditory and visual cues.
\newblock In \emph{Advances in Multimedia Information Processing-PCM 2008},
  pages 317--326. Springer, 2008.

\bibitem[Hassner et~al.(2012)Hassner, Itcher, and Kliper-Gross]{hassner2012}
T.~Hassner, Y.~Itcher, and O.~Kliper-Gross.
\newblock Violent flows: Real-time detection of violent crowd behavior.
\newblock In \emph{Computer Vision and Pattern Recognition Workshops (CVPRW),
  2012 IEEE Computer Society Conference on}, pages 1--6. IEEE, 2012.

\bibitem[Hidaka(2012)]{hidaka2012}
S.~Hidaka.
\newblock Identifying kinematic cues for action style recognition.
\newblock Cognitive Science Society, 2012.

\bibitem[Hofmann(2001)]{hofmann2001}
T.~Hofmann.
\newblock Unsupervised learning by probabilistic latent semantic analysis.
\newblock \emph{Machine learning}, 42\penalty0 (1-2):\penalty0 177--196, 2001.

\bibitem[Huesmann and Eron(2013)]{huesmann2013}
L.~R. Huesmann and L.~D. Eron.
\newblock \emph{Television and the aggressive child: A cross-national
  comparison}.
\newblock Routledge, 2013.

\bibitem[Huesmann and Taylor(2006)]{huesmann2006}
L.~R. Huesmann and L.~D. Taylor.
\newblock The role of media violence in violent behavior.
\newblock \emph{Annu. Rev. Public Health}, 27:\penalty0 393--415, 2006.

\bibitem[Jiang et~al.(2012{\natexlab{a}})Jiang, Dai, Tan, Xue, and
  Ngo]{jiang2012}
Y.-G. Jiang, Q.~Dai, C.~C. Tan, X.~Xue, and C.-W. Ngo.
\newblock The shanghai-hongkong team at mediaeval2012: Violent scene detection
  using trajectory-based features.
\newblock In \emph{MediaEval}, 2012{\natexlab{a}}.

\bibitem[Jiang et~al.(2012{\natexlab{b}})Jiang, Dai, Xue, Liu, and
  Ngo]{jiang2012_1}
Y.-G. Jiang, Q.~Dai, X.~Xue, W.~Liu, and C.-W. Ngo.
\newblock Trajectory-based modeling of human actions with motion reference
  points.
\newblock In \emph{Computer Vision--ECCV 2012}, pages 425--438. Springer,
  2012{\natexlab{b}}.

\bibitem[Jones and Rehg(2002)]{jones2002}
M.~J. Jones and J.~M. Rehg.
\newblock Statistical color models with application to skin detection.
\newblock \emph{International Journal of Computer Vision}, 46\penalty0
  (1):\penalty0 81--96, 2002.

\bibitem[Lam et~al.(2013)Lam, Le, Phan, Satoh, Duong, and Ngo]{lam2013}
V.~Lam, D.-D. Le, S.~Phan, S.~Satoh, D.~A. Duong, and T.~D. Ngo.
\newblock Evaluation of low-level features for detecting violent scenes in
  videos.
\newblock In \emph{Soft Computing and Pattern Recognition (SoCPaR), 2013
  International Conference of}, pages 213--218. IEEE, 2013.

\bibitem[Laptev(2005)]{laptev2005}
I.~Laptev.
\newblock On space-time interest points.
\newblock \emph{International Journal of Computer Vision}, 64\penalty0
  (2-3):\penalty0 107--123, 2005.

\bibitem[Lin and Wang(2009)]{lin2009}
J.~Lin and W.~Wang.
\newblock Weakly-supervised violence detection in movies with audio and video
  based co-training.
\newblock In \emph{Advances in Multimedia Information Processing-PCM 2009},
  pages 930--935. Springer, 2009.

\bibitem[Lowe(2004)]{lowe2004}
D.~G. Lowe.
\newblock Distinctive image features from scale-invariant keypoints.
\newblock \emph{International journal of computer vision}, 60\penalty0
  (2):\penalty0 91--110, 2004.

\bibitem[Mitchell et~al.(2003)Mitchell, Finkelhor, and Wolak]{mitchell2003}
K.~J. Mitchell, D.~Finkelhor, and J.~Wolak.
\newblock The exposure of youth to unwanted sexual material on the internet a
  national survey of risk, impact, and prevention.
\newblock \emph{Youth \& Society}, 34\penalty0 (3):\penalty0 330--358, 2003.

\bibitem[Nam et~al.(1998)Nam, Alghoniemy, and Tewfik]{nam1998}
J.~Nam, M.~Alghoniemy, and A.~H. Tewfik.
\newblock Audio-visual content-based violent scene characterization.
\newblock In \emph{Image Processing, 1998. ICIP 98. Proceedings. 1998
  International Conference on}, volume~1, pages 353--357. IEEE, 1998.

\bibitem[Nievas et~al.(2011)Nievas, Suarez, Garc{\'\i}a, and
  Sukthankar]{nievas2011}
E.~B. Nievas, O.~D. Suarez, G.~B. Garc{\'\i}a, and R.~Sukthankar.
\newblock Violence detection in video using computer vision techniques.
\newblock In \emph{Computer Analysis of Images and Patterns}, pages 332--339.
  Springer, 2011.

\bibitem[OpticalFlow(2015)]{opticalflowimpl}
D.~OpticalFlow.
\newblock Optical flow implementation, 2015.
\newblock URL
  \url{http://docs.opencv.org/modules/video/doc/motion_analysis_and_object_tracking.html#calcopticalflowfarneback}.
\newblock Online: accessed 21-Oct-2015.

\bibitem[Parker(2011)]{parker2011}
C.~Parker.
\newblock An analysis of performance measures for binary classifiers.
\newblock In \emph{Data Mining (ICDM), 2011 IEEE 11th International Conference
  on}, pages 517--526. IEEE, 2011.

\bibitem[Pedregosa et~al.(2011)Pedregosa, Varoquaux, Gramfort, Michel, Thirion,
  Grisel, Blondel, Prettenhofer, Weiss, Dubourg, Vanderplas, Passos,
  Cournapeau, Brucher, Perrot, and Duchesnay]{scikit-learn}
F.~Pedregosa, G.~Varoquaux, A.~Gramfort, V.~Michel, B.~Thirion, O.~Grisel,
  M.~Blondel, P.~Prettenhofer, R.~Weiss, V.~Dubourg, J.~Vanderplas, A.~Passos,
  D.~Cournapeau, M.~Brucher, M.~Perrot, and E.~Duchesnay.
\newblock Scikit-learn: Machine learning in {P}ython.
\newblock \emph{Journal of Machine Learning Research}, 12:\penalty0 2825--2830,
  2011.

\bibitem[Platt et~al.(1999)]{platt1999}
J.~Platt et~al.
\newblock Probabilistic outputs for support vector machines and comparisons to
  regularized likelihood methods.
\newblock \emph{Advances in large margin classifiers}, 10\penalty0
  (3):\penalty0 61--74, 1999.

\bibitem[Pogrebnyak et~al.(2015)Pogrebnyak, Timoshenko, Burcev, and
  Kulinkin]{pogrebnyak2015}
M.~Pogrebnyak, D.~Timoshenko, I.~Burcev, and A.~Kulinkin.
\newblock Adult-content detection in video with the use of nvidia gpu.
\newblock 2015.

\bibitem[Richardson(2013)]{richardson2013}
L.~Richardson.
\newblock Beautiful soup.
\newblock \emph{Crummy: The Site}, 2013.
\newblock URL \url{http://www.crummy.com/software/BeautifulSoup/}.

\bibitem[Rijsbergen(1979)]{Rijsbergen1979}
C.~J.~V. Rijsbergen.
\newblock \emph{Information Retrieval}.
\newblock Butterworth-Heinemann, Newton, MA, USA, 2nd edition, 1979.
\newblock ISBN 0408709294.

\bibitem[Saerbeck and Bartneck(2010)]{saerbeck2010}
M.~Saerbeck and C.~Bartneck.
\newblock Perception of affect elicited by robot motion.
\newblock In \emph{Proceedings of the 5th ACM/IEEE international conference on
  Human-robot interaction}, pages 53--60. IEEE Press, 2010.

\bibitem[Schedl et~al.()Schedl, Sj{\"o}berg, Mironica, Ionescu, Quang, and
  Jiang]{schedl6vsd2014}
M.~Schedl, M.~Sj{\"o}berg, I.~Mironica, B.~Ionescu, V.~L. Quang, and Y.-G.
  Jiang.
\newblock Vsd2014: A dataset for violent scenes detection in hollywood movies
  and web videos.
\newblock \emph{Sixth Sense}, 6\penalty0 (2.00):\penalty0 12--40.

\bibitem[Schulze et~al.(2014)Schulze, Henter, Borth, and Dengel]{schulze2014}
C.~Schulze, D.~Henter, D.~Borth, and A.~Dengel.
\newblock Automatic detection of csa media by multi-modal feature fusion for
  law enforcement support.
\newblock In \emph{Proceedings of International Conference on Multimedia
  Retrieval}, page 353. ACM, 2014.

\bibitem[Sokolova and Lapalme(2009)]{sokolova2009}
M.~Sokolova and G.~Lapalme.
\newblock A systematic analysis of performance measures for classification
  tasks.
\newblock \emph{Information Processing \& Management}, 45\penalty0
  (4):\penalty0 427--437, 2009.

\bibitem[Sparks(2015)]{sparks2015}
G.~Sparks.
\newblock \emph{Media effects research: A basic overview}.
\newblock Cengage Learning, 2015.

\bibitem[Tompkins(2003)]{tompkins2003}
A.~Tompkins.
\newblock The psychological effects of violent media on children.
\newblock \emph{AllPsych Journal}, 14, 2003.

\bibitem[Wesch(2008)]{youtubestats}
M.~Wesch.
\newblock Youtube statistics, 2008.
\newblock URL \url{http://mediatedcultures.net/thoughts/youtube-statistics/}.
\newblock Online: accessed 08-Jan-2016.

\bibitem[Wikipedia(2015)]{wiki:opticalflow}
Wikipedia.
\newblock Optical flow, 2015.
\newblock URL \url{https://en.wikipedia.org/wiki/Optical_flow}.
\newblock Online: accessed 21-Oct-2015.

\end{thebibliography}

\end{document}